\documentclass[10pt,twocolumn,letterpaper]{article}

\usepackage{cvpr}
\usepackage{times}
\usepackage{epsfig}
\usepackage{graphicx}
\usepackage{amsmath}
\usepackage{amssymb}
\usepackage{color}
\usepackage{cite}
\usepackage{dsfont}
\usepackage{enumitem}
\usepackage{subfigure}

\usepackage[breaklinks=true,bookmarks=false]{hyperref}

\cvprfinalcopy %

\ifcvprfinal\fi
\begin{document}

\title{ECO: Egocentric Cognitive Mapping}

\author{Jayant Sharma\\
University of Minnesota\\
{\tt\small sharm546@umn.edu}
\and
Zixing Wang\\
University of Minnesota\\
{\tt\small wang7923@umn.edu}
\and
Alberto Speranzon\\
Honeywell\\
{\tt\small Alberto.Speranzon@honeywell.com}
\and
Vijay Venkataraman\\
Honeywell\\
{\tt\small Vijay.Venkataraman@honeywell.com}
\and
Hyun Soo Park\\
University of Minnesota\\
{\tt\small hspark@umn.edu}
}

\maketitle

\begin{abstract}
We present a new method to localize a camera within a previously unseen environment perceived from an egocentric point of view. Although this is, in general, an ill-posed problem, humans can effortlessly and efficiently determine their relative location and orientation and navigate into previously unseen environments, e.g., finding a specific item in a new grocery store. To enable such capability, we design a new egocentric representation, which we call ECO (Egocentric COgnitive map). ECO is biologically inspired, by the cognitive map that allows human navigation, and it encodes the surrounding visual semantics with respect to both distance and orientation. ECO has three main properties: (1) reconfigurability: complex semantics and geometry is captured via the synthesis of atomic visual representations; %
(2) robustness: the visual semantics are registered in a geometrically consistent way,
enabling us to learn meaningful atomic representations; (3) adaptability: a domain adaptation framework is designed to generalize the learned representation without manual calibration. As a proof-of-concept, we use ECO to localize a camera within real-world scenes---various grocery stores---and demonstrate performance improvements when compared to existing semantic localization approaches.

\begin{figure}
  \centering  
      \subfigure[CNN features]{\label{fig:pca_cnn}\includegraphics[width=1\columnwidth]{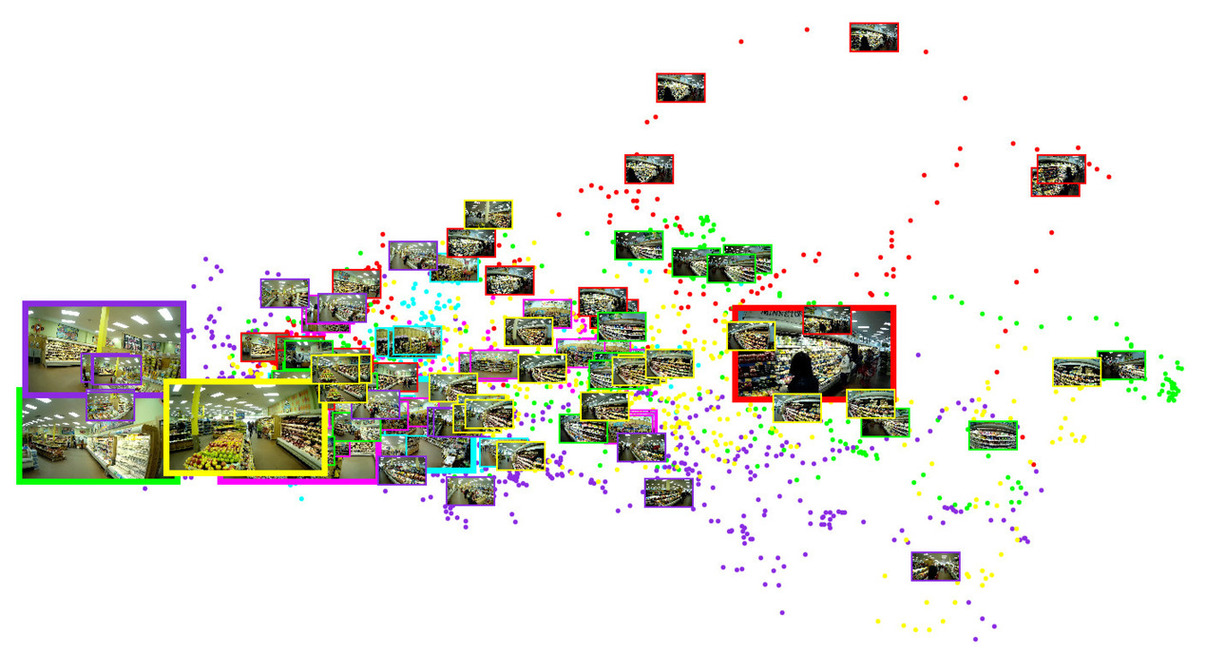}}~
      \vspace{-3mm}
      
      \subfigure[ECO features]{\label{fig:pca_box}\includegraphics[width=1\columnwidth]{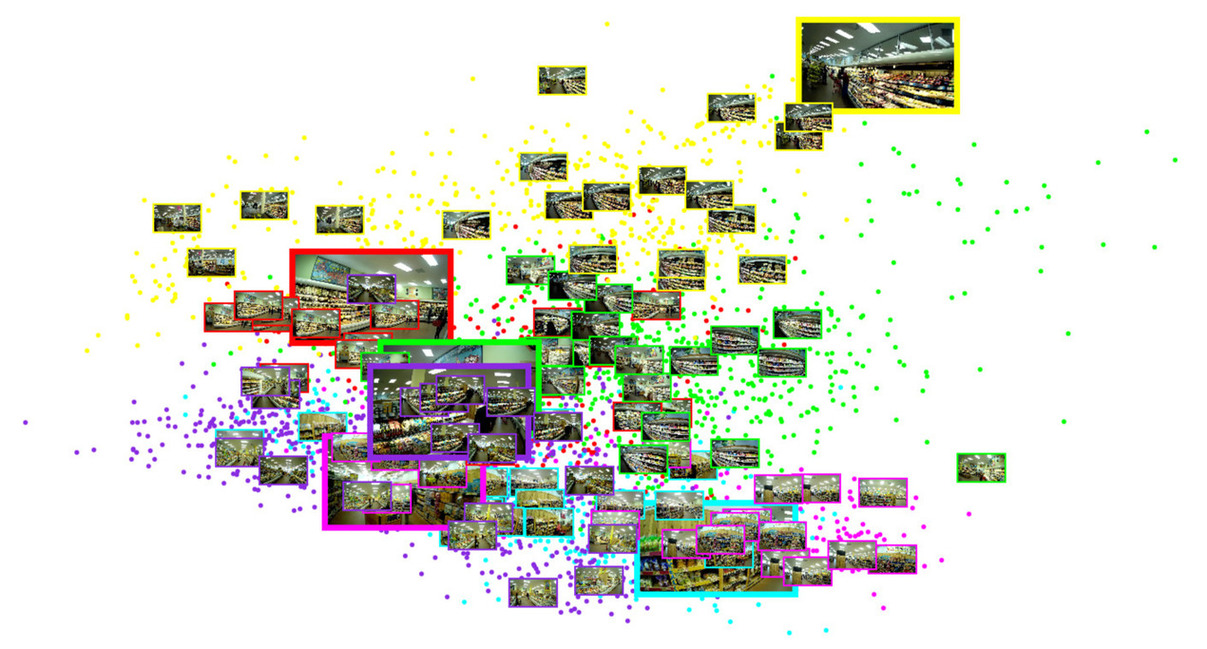}}~
  \caption{CNN vs ECO feature embeddings of six different semantic categories from three grocery stores using PCA. ECO representation is well-aligned, robust to geometry and adaptable across different stores, while CNN representation is noisy and non-discriminative.} \label{Fig:warping}
\end{figure}

\end{abstract}

\section{Introduction}
Imagine you are visiting a large grocery store that you have never been to before. You see the dairy items on the left shelf and the seasoning corner ahead of you. Everything you see is new, yet you are still able to orient yourself and effortlessly navigate in the environment. %

Humans have developed the capability of building what is called a  {\em cognitive map}---a mental representation to perceive, store, and recall relative location by learning semantic and geometric information~\cite{Tolman:1948}. Such cognitive map is both robust and reconfigurable: it provides a sense of direction and distance in the semantic world. Humans can readily construct a new representation by reconfiguring their past memories\footnote{Such ability has been called the {\em GPS of the brain} and the hippocampus is known to be responsible to construct the cognitive map~\cite{okeefe:1996}.}. For instance, you can localize yourself in a new store despite the fact it has a new spatial layout and visual patterns. We can do this, without accurate spatial information or photographic memory (e.g., exact item label, spatial arrangement of items, etc.). The ability to build such a cognitive map from (current and past) structured and semantically rich data, will enable intelligent systems, such as mobile robots, to explore and navigate uncharted spaces. 

Inspired by the cognitive map, in this paper, we present a new representation, called {\em ECO} (Egocentric COgnitive Map), built from a first person video. ECO has the following properties: (1) Reconfigurability: the cognitive map is constructed sequentially via temporal saccade movements related to the retinal structure of human eyes, i.e., the representation of each region in the field of view is reconstructed by parts. %
We design the map by assembling visual semantic data and use the spatial layout of the new scene as a geometric constraint; (2) Robustness: analogous to the human ability of mental re-scaling~\cite{Wilma:2015}, visual semantics need to be stabilized using 3D geometric representations (e.g., vanishing point, gravity, and spatial layout)~\cite{park_cvpr_future_loc:2016,kopf:2014}. Further, as the visual scene around the camera is distorted with respect to an egocentric coordinate (e.g., a farther object appears smaller~\cite{ladicky:2014}) this results in distinctive representations for the same object; (3) Adaptability: albeit sharing similar layout and visual patterns, different scenes never look the same, e.g., other people in the scene, illumination, and item arrangements make such scene quite unique. To build a generalizable cognitive map, the representation needs to be transformable, so that it can be adapted to a new scene without explicitly specifying scene correspondences. 

Existing visual localization frameworks in computer vision~\cite{baatz:2012,torralba:2003,hays:2008,doersch2012what,kendall2017geometric,kendall2015convolutional,frahm:2010,snavely:2006,CVPR14_Khosla}, lack one or more of these properties. Proposed representations are, mostly, built either upon low level geometric features (e.g., camera resectioning), which are reconfigurable but often fragile when faced with a new scene, or high level semantic features, which are more robust but not reconfigurable (e.g., depend on specific classifications). In the spectrum of existing localization methods, ECO is situated in the middle, as it preserves both relative geometry (e.g. middle of aisle) and visual semantics (e.g. item categories).

Our system computes the ECO representation by taking, as input, an egocentric image and, by retrieving images from the training data, it computes the relative localization in a new space. ECO is constructed by a linear combination of visual semantic information, associated to atomic object-centric image patches. We estimate the scene geometry to identify the spatial layout and transform the egocentric image patches to a canonical scale and orientation, where visual semantics are efficiently learned. An adversarial domain adaptation approach is used to transfer to a new scene. 

The key contributions of the paper include: (1) a new egocentric representation that is reconfigurable, robust, and adaptable, which enables localization in a new scene; (2) a novel geometric transformation to rectify an egocentric image to an object-centric coordinate frame; (3) a new first-person grocery dataset that allows us to explore unseen localization and navigation; and (4) performance evaluation of ECO showing we can accurately identify scenes in the presence of a large variation of spatial layouts,
without the need of fine tuning.

\noindent\textbf{Challenges of First-person Videos} While a first-person video directly taps into the wearer's visual sensation, building a persistent representation for first-person videos is still challenging. Egocentric videos are highly dynamic, local (limited field of view), and person-biased due to severe head movements, which generates a larger variation of visual data. Due to such variations, learning a robust representation without a proper stabilization is intractable, requiring massive amount of annotated data. Our new representation leverages geometric features to register visual semantic information in 3D, which is invariant to head movement, making the learning feasible.

\begin{figure*}[t]
  \centering  
      \subfigure[Cheese at store 1]{\label{fig:cheese_2}\includegraphics[width=0.3\textwidth]{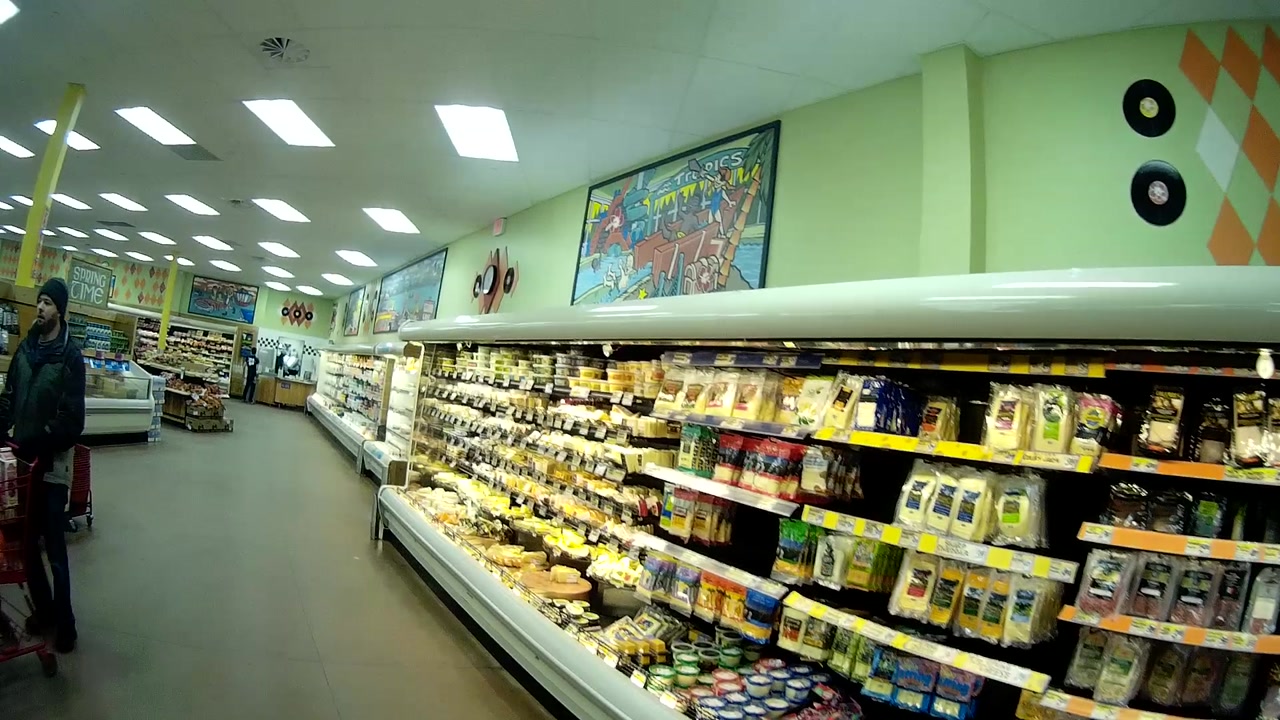}}~
      \subfigure[Cheese at store 2]{\label{fig:cheese_1}\includegraphics[width=0.3\textwidth]{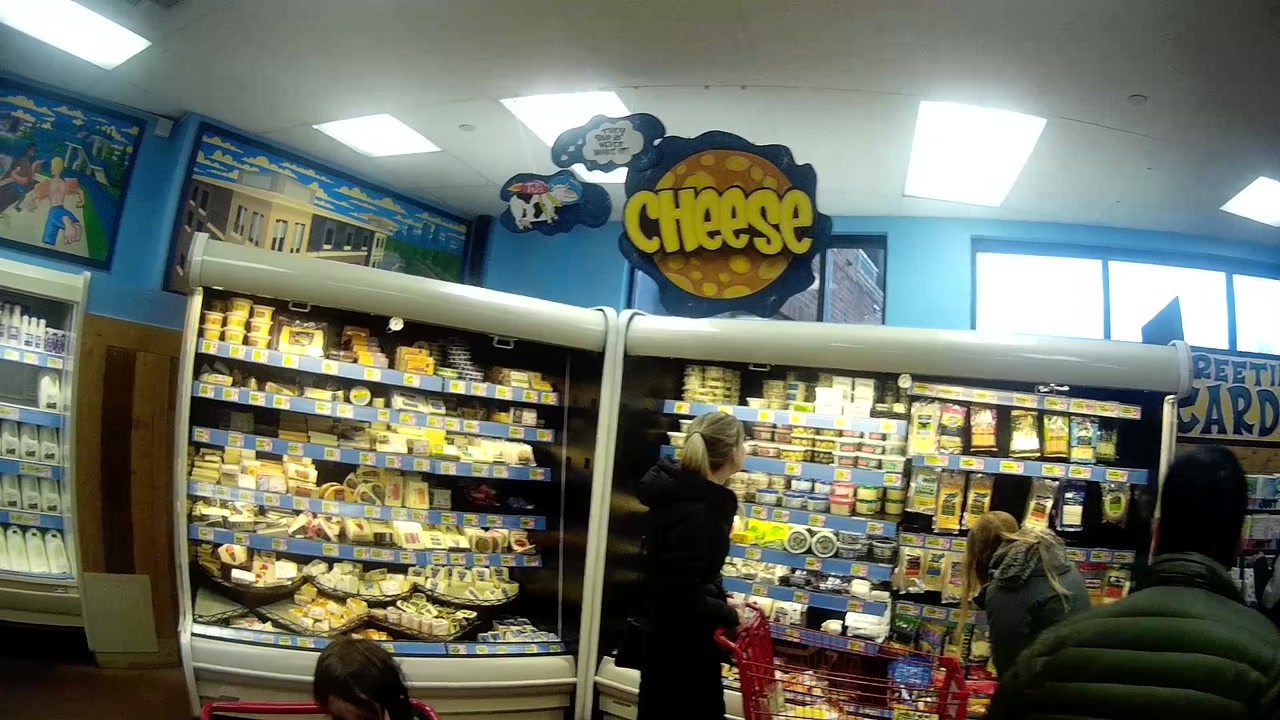}}~
      \subfigure[Dairy at store 1]{\label{fig:dairy_1}\includegraphics[width=0.3\textwidth]{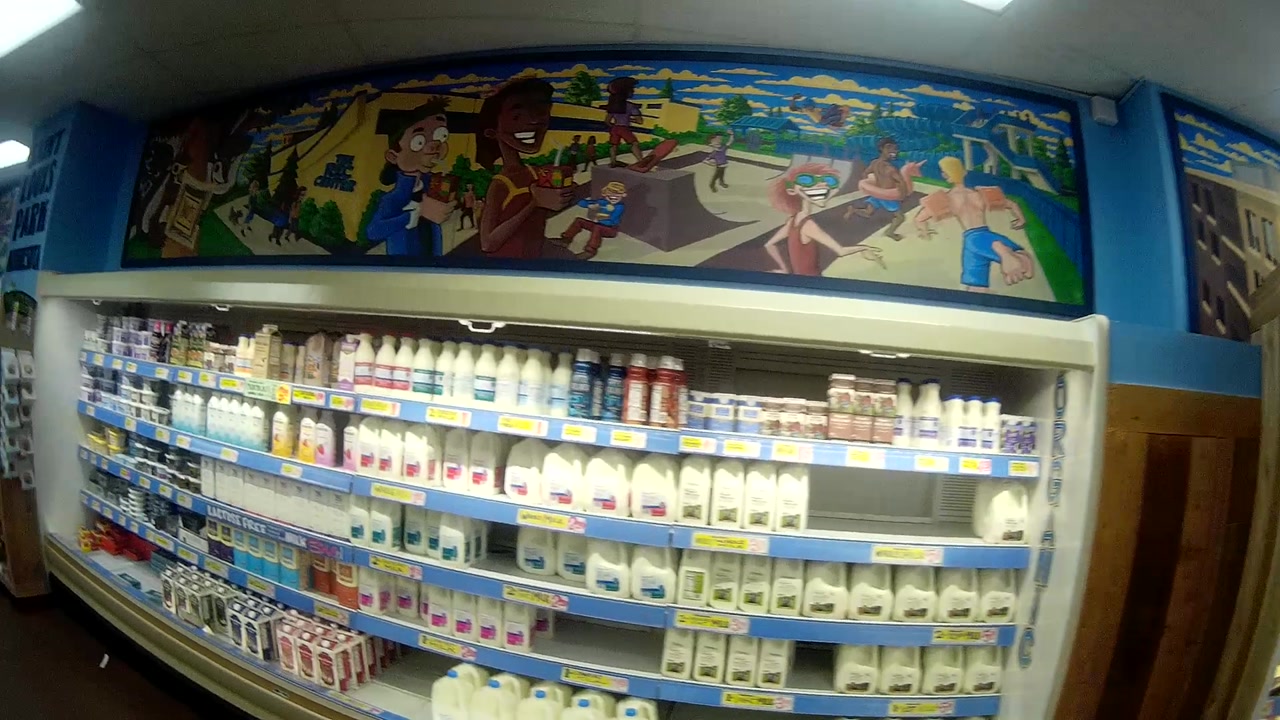}}
  \caption{The images of cheese sections from two stores (a,b) are highly distinctive in terms of spatial layout and appearance. A global scene descriptors are sensitive to the scene layout, which often matches to the image of dairy section (a,c). } \label{Fig:warping}
\end{figure*}

\section{Related Work}
The main goal of this paper is to learn a persistent visual representation for first-person videos that preserves relevant spatial context. In this section, we briefly review the visual localization process that approximates the cognitive map, and the first-person representation.

\subsection{Visual Localization} Image localization techniques often incorporate other correlated sensory data. Cozman and Krotkov~\cite{cozman:1995} introduced localization of an image taken from unknown environment using temporal changes in sun altitudes. Jacobs et al.~\cite{jacobs:2007} incorporated weather data reported by satellite imagery to localize widely distributed cameras. They find matches between weather conditions on images over an year and the expected weather changes indicated by satellite imagery. As GPS became a viable solution for localization in many applications, GPS-tagged images can help to localize images that do not have such tags. Zhang and Kosecka~\cite{zhang:2006} built a GPS-tagged image repository in urban environments and found correspondences between a query image and the database using SIFT features~\cite{lowe:2004}. Hays and Efros~\cite{hays:2008} leveraged GPS-tagged internet images to estimate a location probability distribution over Earth. Kalogerakis et al.~\cite{kalogerakis:2009} extended the work to disambiguate locations of the images without distinct landmarks. %
Baatz et al.~\cite{baatz:2012} estimated image location based on a 3D elevation model of mountainous terrains and evaluated their method on the scale of a country (Switzerland). 

Two main approaches have been used for image based localization. (1) Recognition based localization: Torralba et al.~\cite{torralba:2003} used global context to recognize a scene category using a hidden Markov model framework. Se et al.~\cite{se:2002} applied a RANSAC framework for global camera registration. Robertson and Cipolla~\cite{robertson:2004} estimated image positions relative to a set of rectified views of building facades registered onto a city map. Recently, deep neural networks with a large-scale data continue to push the boundary of localization performance to human level~\cite{CVPR14_Khosla,doersch2012what,kendall2017geometric,kendall2015convolutional}. (2) Camera resectioning based localization: Structure from motion have also been employed for large scale image localization. Snavely et al.~\cite{snavely:2006} exploited structure from motion to browse a photo collection from the exact location where it was taken. They used hundreds of images for pose registration in 3D. Agarwal et al.~\cite{agarwal:2009} presented a parallelizable system that can reconstruct hundreds of thousands of images (city scale) within two days. Frahm et al~\cite{frahm:2010} showed larger scale reconstruction (millions of images) that can be executed on a single PC. Unlike existing visual localization frameworks that rely on geometry or visual semantics, our cognitive map representation ties geometry and visual semantics to build a robust first-person representation that can be reliably matched to the relevant spatial context.

\subsection{First-person Representation} A first-person camera sees what the camera wearer sees, which differs from a third person system such as surveillance cameras. This enables us to measure subtle head movement, information which is used in behavioral science to study quality of life and develop relevant technology~\cite{kanade:2012, rehg:2013, pusiol:2014}. This has motivated vision tasks such as understanding fixation point~\cite{li:2013}, identifying eye contact~\cite{ye:2015}, and localizing joint attention~\cite{fathi:2012,park:2012}. 

A first-person camera ego-motion is a highly discriminative feature for activity recognition. 
Fathi et al.~\cite{fathi:2011, fathi:2012} used gaze and object segmentation cues to classify activities. 2D motion features were exploited by Kitani et al.~\cite{kitani:2011} to categorize and segment a first-person sport video in a unsupervised manner. Coarse-to-fine motion models~\cite{ryoo:2013} and a pretrained convolutional neural network~\cite{ryoo:2015} provided a strong cue to recognize activities. Yonetani et al.~\cite{yonetani:2015} utilized a motion correlation between first and third person videos to recognize people's identity. Kopf et al.~\cite{kopf:2014} stabilized first-person footage via 3D reconstruction of camera ego-motion. In a social setting, joint attention was estimated via triangulation of multiple camera optical rays~\cite{park:2012,park:2015} and the estimated joint attention was used to edit social video footage~\cite{arev:2014}. Another information that the first-person camera captures is exomotion or scene motion. Pirsiavash and Ramanan~\cite{pirsiavash:2012} used an object centric representation and temporal correlation to recognize active/passive objects from a egocentric video, and Rogez et al.~\cite{rogez:2015} leveraged a prior distribution of body and hand coordination to estimate poses from a chest mounted RGBD camera. Lee et al.~\cite{lee:2012} summarized a life-logging video by discovering important people and objects based on temporal correlation, and Xiong and Grauman~\cite{xiong:2014} utilized a web image prior to select a set of good images from egocentric videos. Fathi et al.~\cite{fathi:2012} used observed faces to identify social interactions and Pusiol et al.~\cite{pusiol:2014} learned a feature that indicates joint attention in child-caregiver interactions. As the camera wearer interacts with surroundings, first-person videos can encode affordances (understanding on how objects can be used or how the environment can be traversed) from the scene. For instance, semantic meaning of scene 3D layout (e.g., building, road, and street signs) tells us about motion affordance, allowing predicting future activities~\cite{park_cvpr_future_loc:2016,singh:2016} and objects to interact~\cite{bertasius:2016}. Gaze direction can be precisely estimated from visual semantics and 3D motion of first-person videos~\cite{li:2013}, important objects can be detected~\cite{pirsiavash:2012, lee:2012}, visual transformation can be predicted through ego-motion~\cite{jayaraman:2015}, and robust feature can be learned~\cite{agrawal:2015}.

\section{Egocentric Cognitive Map (ECO)}
In the following we formalize and describe in detail the key properties of ECO.

\subsection{Reconfigurability}\label{sec:reconfigurability}
Given an egocentric image $\mathbf{I}$, we compute a feature map $f(\mathbf{I}) \in \mathds{R}^D$ where $D$ is the feature dimension. This has been studied in scene classification~\cite{torralba:2003,robertson:2004,CVPR14_Khosla,doersch2012what,kendall2017geometric,kendall2015convolutional} to describe a scene with a global descriptor where its expressivity is fundamentally bounded by the number of observed images at the corresponding 3D camera pose. For example, the image of a cheese section in a store, $\mathbf{I}_{\rm c}^1$, in Figure~\ref{fig:cheese_2} looks different than $\mathbf{I}_{\rm c}^2$, see Figure~\ref{fig:cheese_1}, in terms of spatial layout, geometric alignment, perspective distortion and background clutter. It is more similar to the image $\mathbf{I}_{\rm d}$, related to the dairy section, as in Figure~\ref{fig:dairy_1}. Thus we have that $\|f(\mathbf{I}_\mathrm{c}^1)-f(\mathbf{I}_\mathrm{c}^2)\| > \|f(\mathbf{I}_\mathrm{c}^2)-f(\mathbf{I}_\mathrm{d})\|$
where $f$ is the global scene feature descriptor (e.g., CNN feature). The atomic representation of the ECO can alleviate the pitfall of this global descriptors. 

\begin{figure*}[t]
    \centering
    \subfigure{\label{Fig:raw_1}\includegraphics[width=0.24\textwidth]{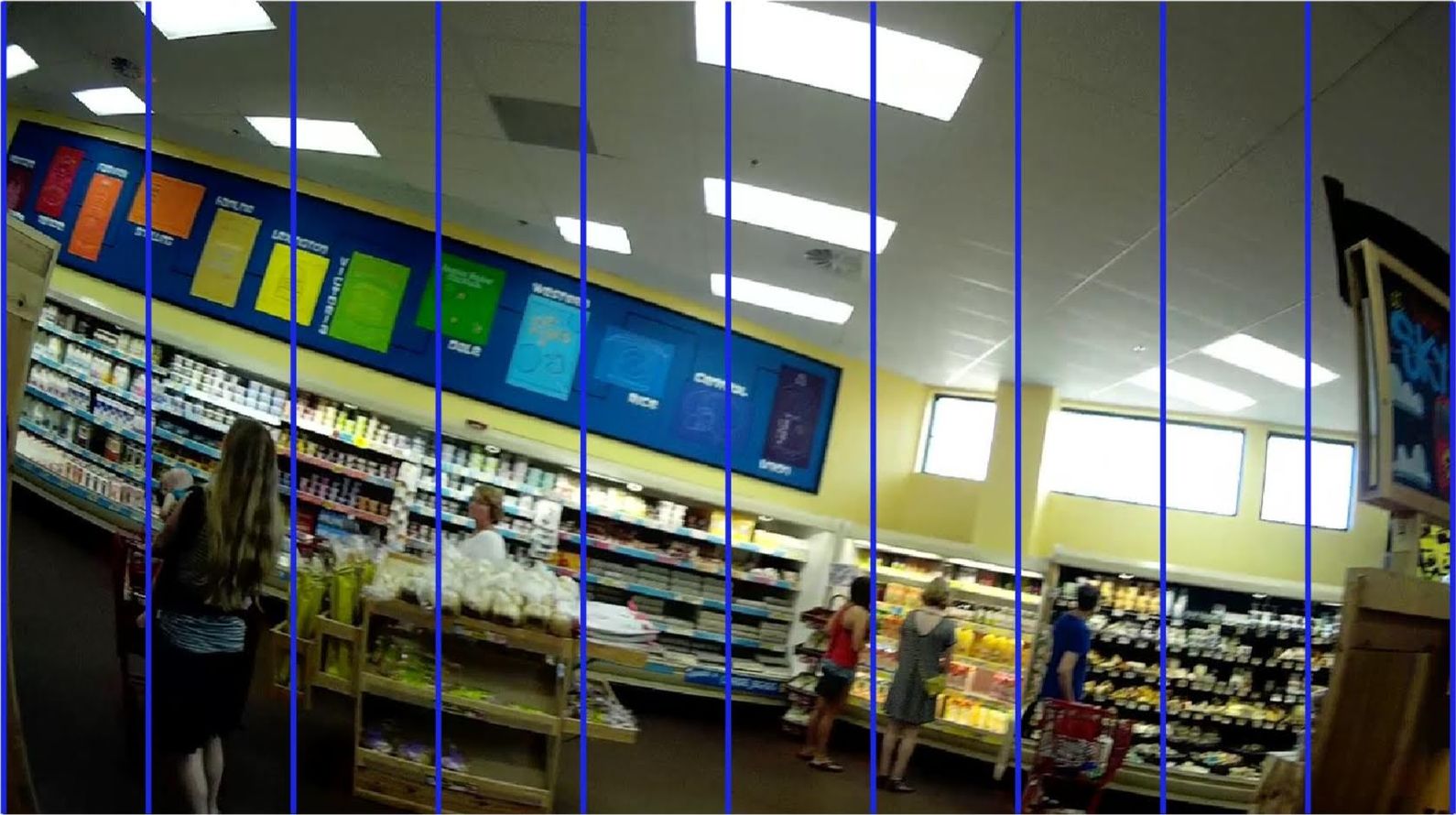}}
    \vspace{-3mm}
    \subfigure{\label{Fig:raw_2}\includegraphics[width=0.24\textwidth]{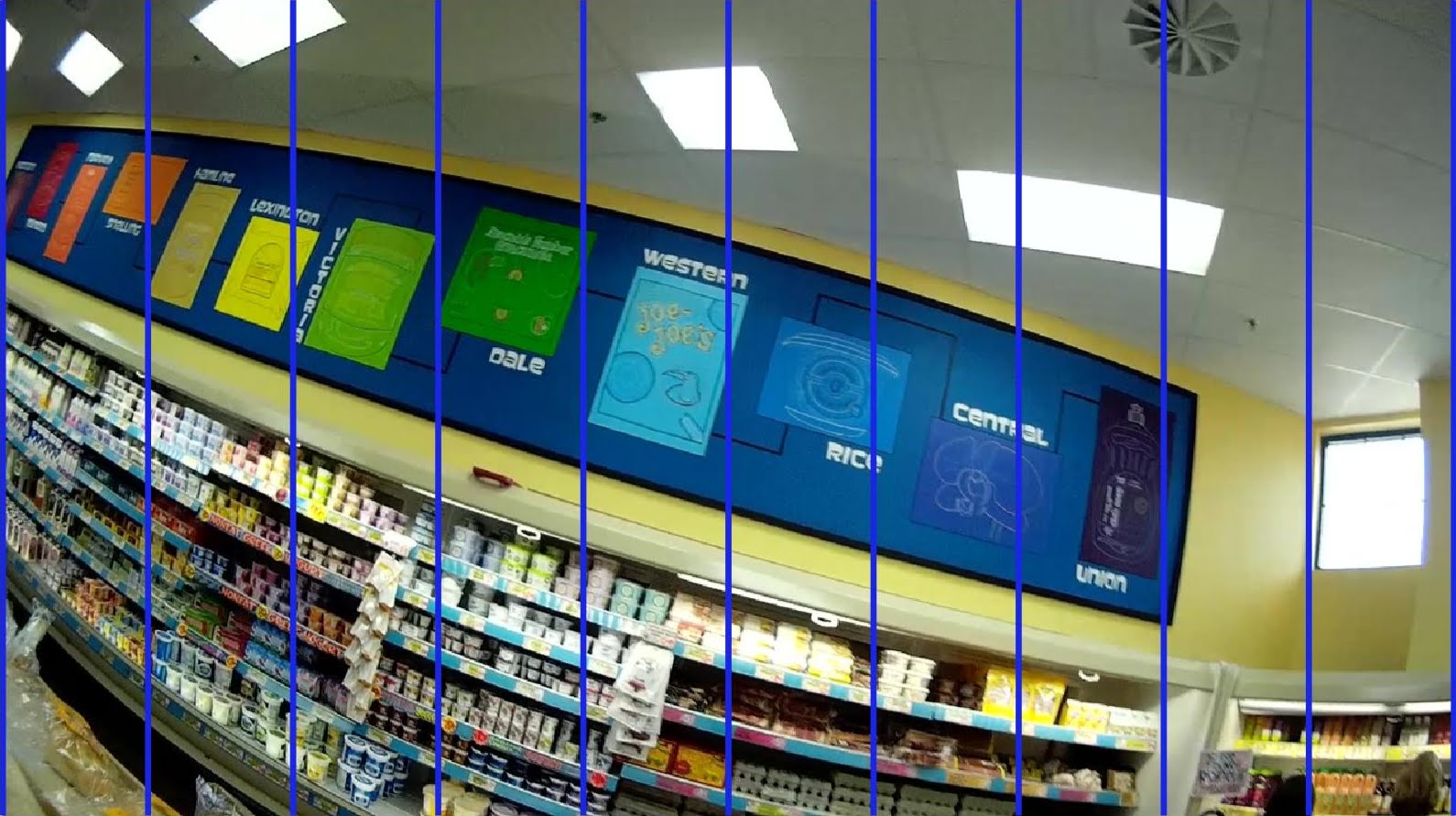}}
    \subfigure{\label{Fig:raw_3}\includegraphics[width=0.24\textwidth]{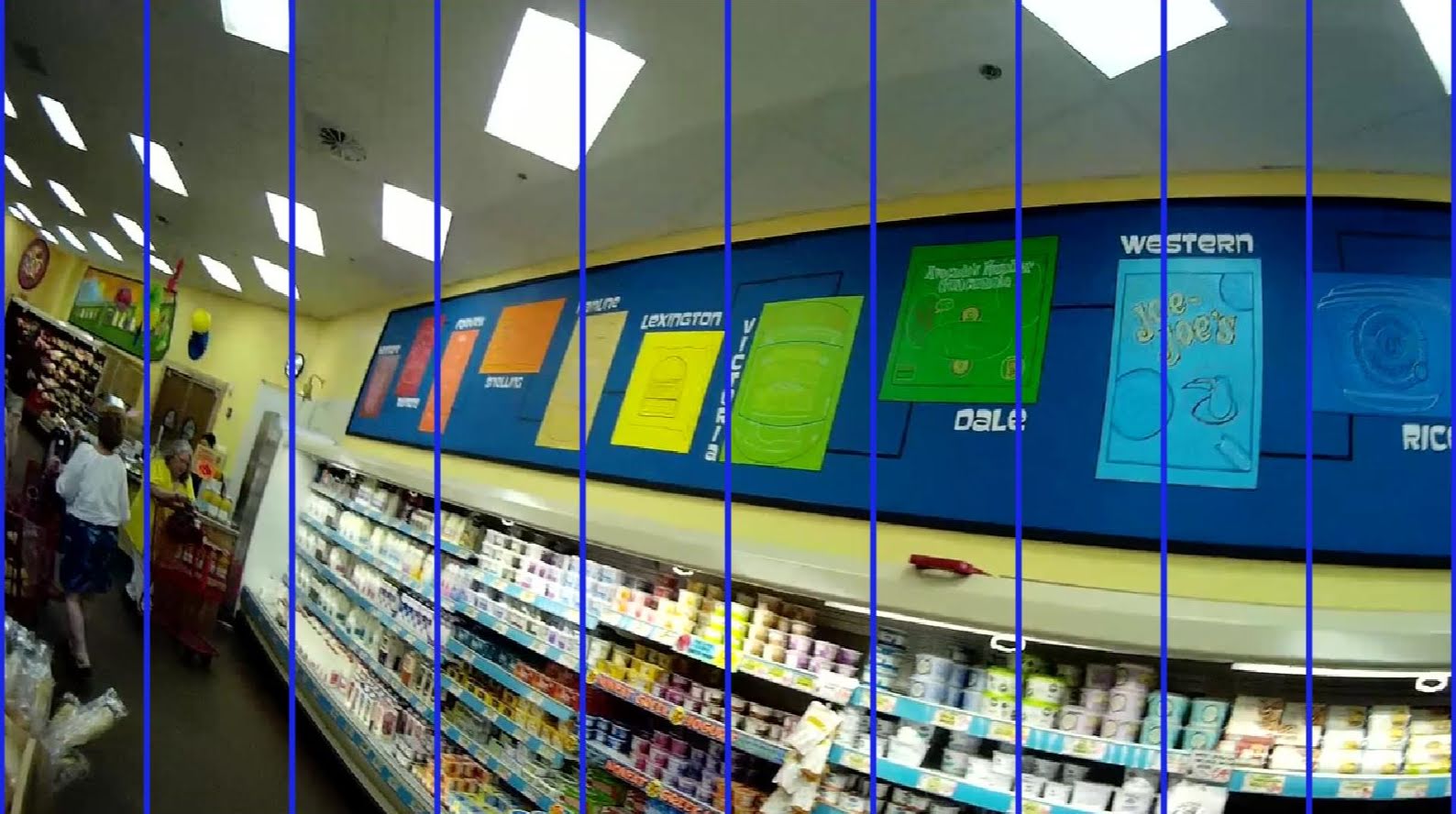}}
    \subfigure{\label{Fig:raw_4}\includegraphics[width=0.24\textwidth]{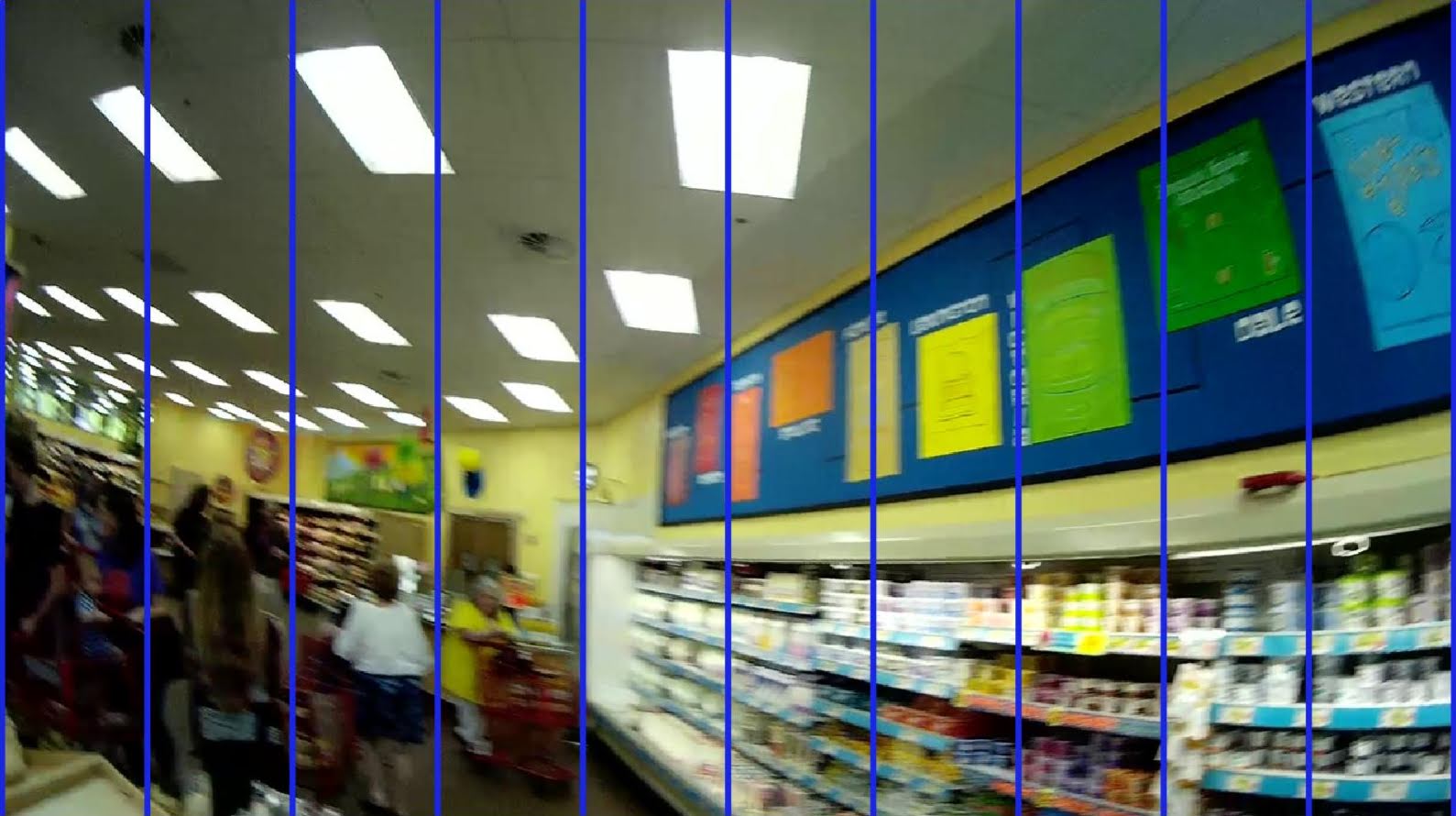}}

    \subfigure{\label{Fig:raw_1}\includegraphics[width=0.24\textwidth]{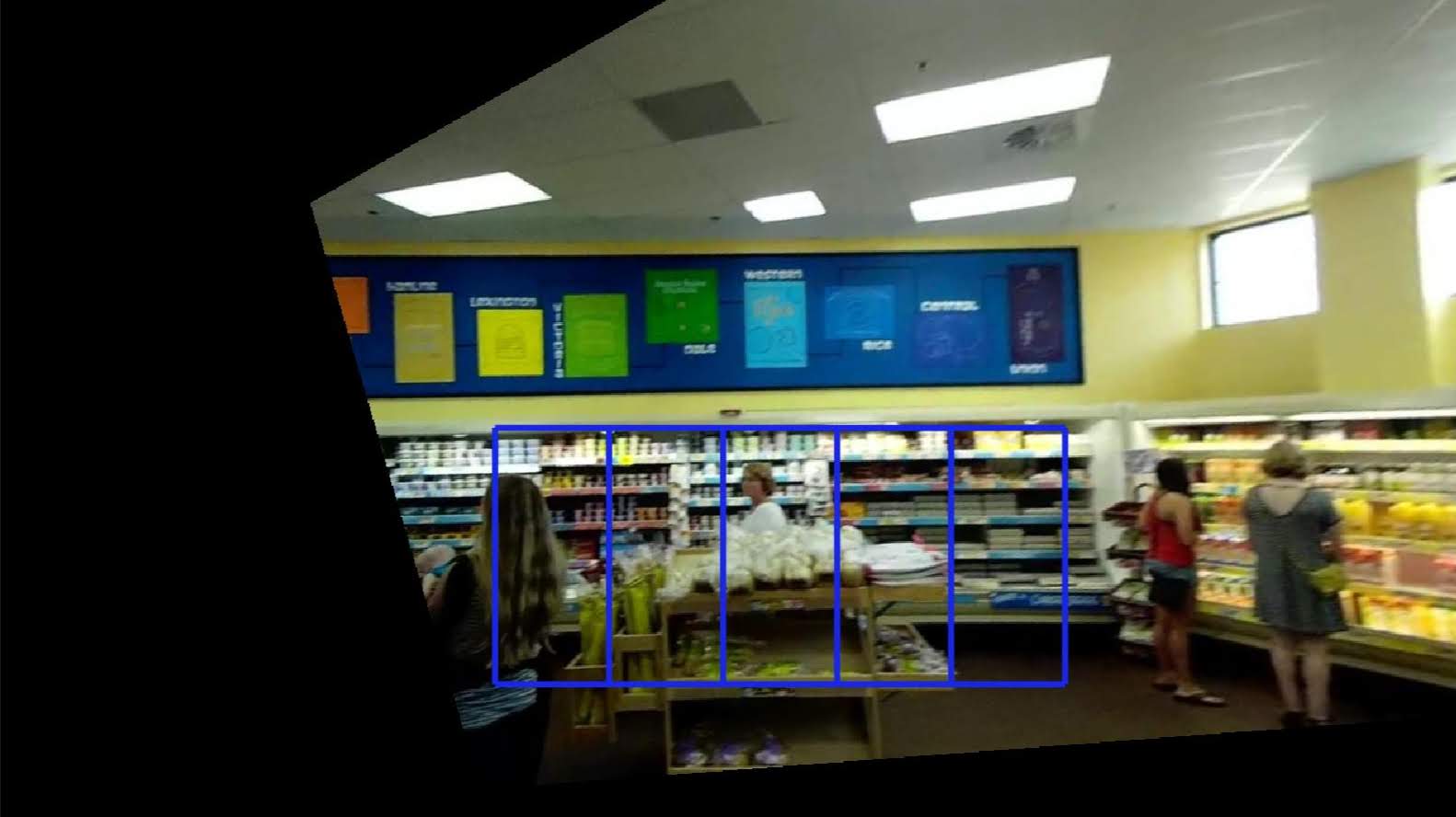}}
    \subfigure{\label{Fig:raw_2}\includegraphics[width=0.24\textwidth]{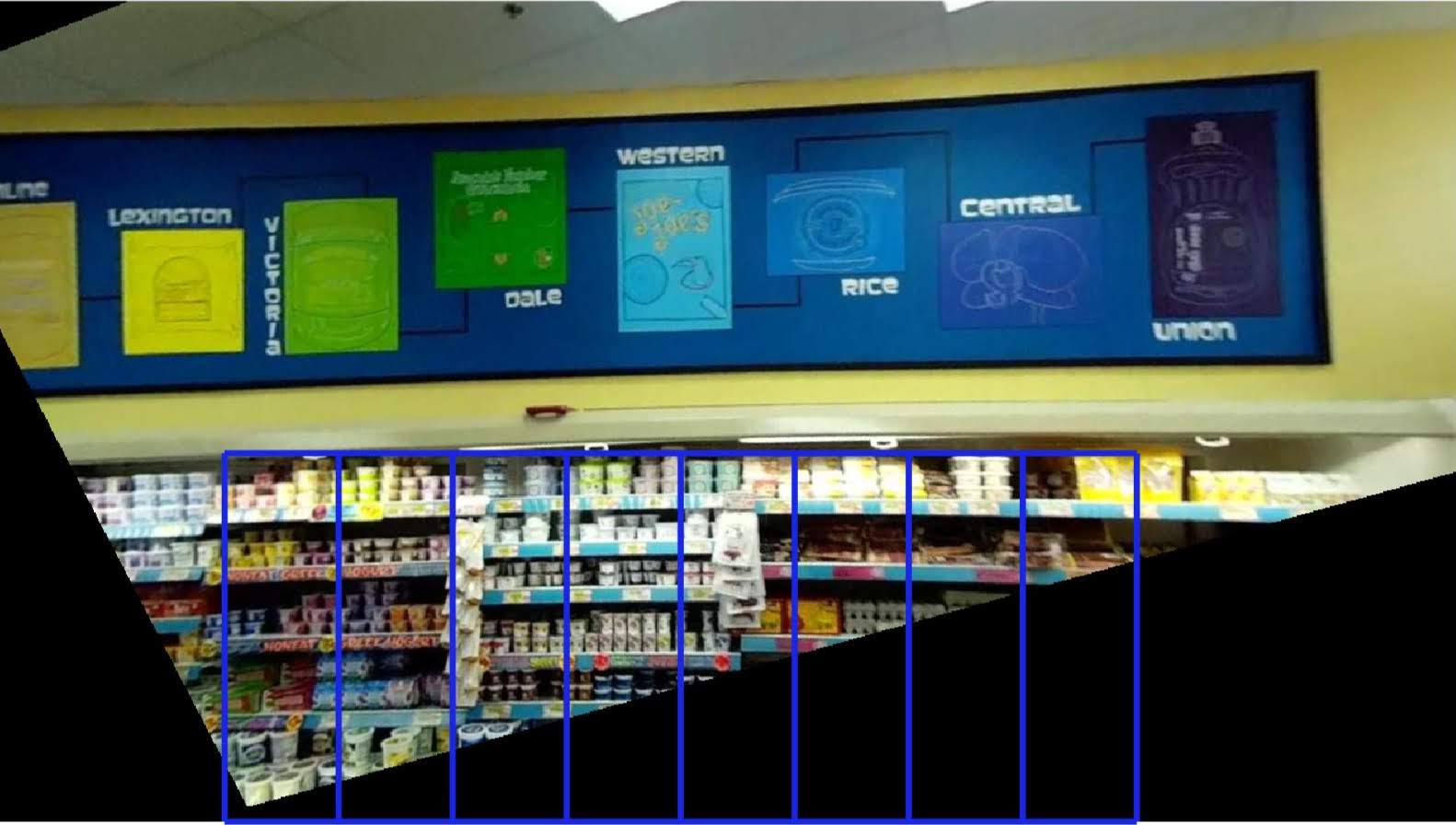}}
    \subfigure{\label{Fig:raw_3}\includegraphics[width=0.24\textwidth]{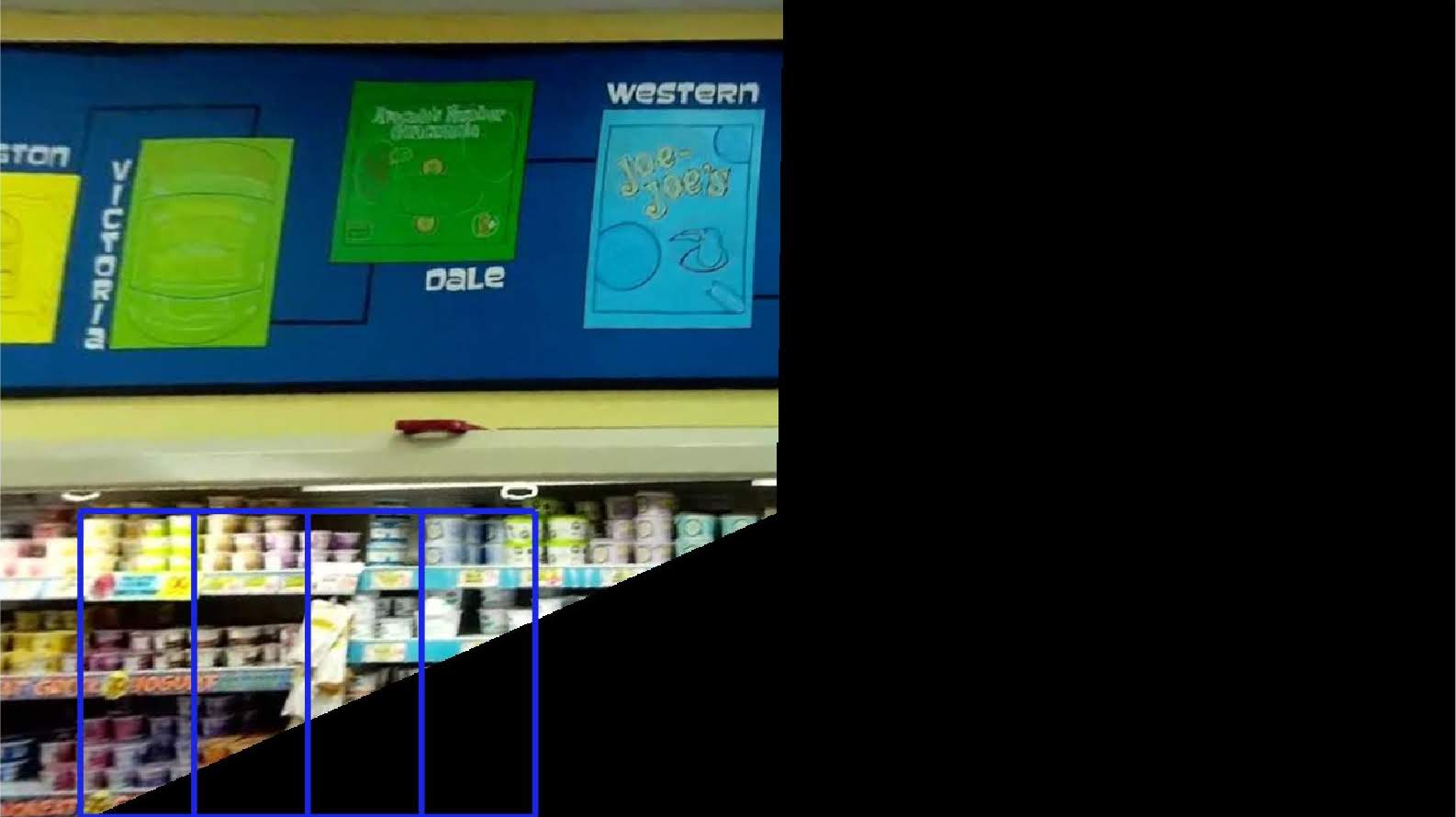}}
    \subfigure{\label{Fig:raw_4}\includegraphics[width=0.24\textwidth]{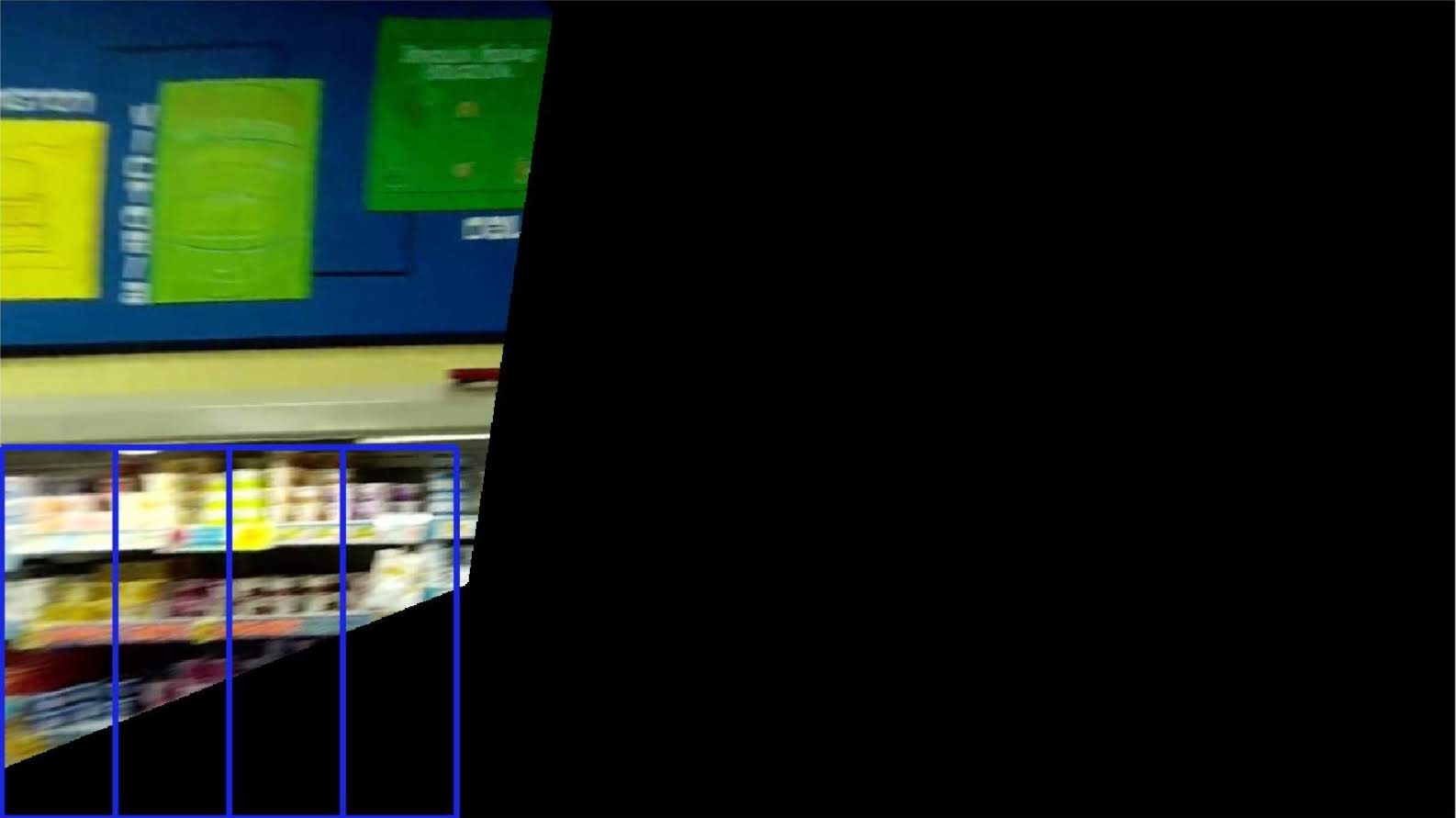}}

\caption{Image Warping -- rows from top to bottom demonstrate (a) raw images, (c) strips from frontal views as the dairy section is approached and passed-by}
\label{Fig:warping}
\vspace{-3mm}
\end{figure*}

However, our surrounding space is rather compositional, i.e., there exists an infinite number of 3D spatial configurations and viewpoints/orientations to form diverse egocentric images. 
Instead, we represent an egocentric image by a set of atomic image patches:
\begin{align}
f(\mathbf{I}) = \frac{1}{\sum_{i=1}^n w_i}\sum_{i=1}^n w_i\overline{f}(\mathbf{I}_i) \label{Eq:reconfigurability}
\end{align}
where $\mathbf{I}_i \subseteq \mathbf{I}$ is the $i$-th image patch, $w_i\in \mathds{R}$ is its weight, and $\overline{f} \in \mathds{R}^D$ maps an atomic image patch to features. This representation is reconfigurable: a scene can be reconstructed by synthesizing a set of image patches, which is resilient to a variation of spatial layouts and camera pose. Therefore, even though the scene geometry (spatial layout) and semantics (visual appearance) in a new space are highly distinctive, it is possible to associate a new scene with a previously observed one. Note that the weight $w_i$ controls the importance of the image patches. For example, the weights can be set to the inverse distance, $w_i = 1/r$ where $r$ is the distance to the corresponding 3D object, analogous to proxemics~\cite{hall:1963}.

\subsection{Robustness}
Humans associate an object with a canonical mental image of an object that is robust to their viewpoint changes, i.e., we can rotate and scale the current visual scene to match to the mental image. On the other hand, a first-person camera produces non-iconic images, i.e., an object appears in a skewed shape as seen from an oblique view. For instance, a sequence of egocentric images containing dairy section in a grocery store from which we wish to learn the visual semantics of the dairy section. Due to the dynamic scenes induced by severe head motion, the dairy section appears in various alignments and grows/shrinks as the camera approaches/moves away. Due to the non-iconic nature of egocentric images, multiple visual representations map to a single object. To address this, we modify Equation~(\ref{Eq:reconfigurability}) such that
\begin{equation}
    f(\mathbf{I}) = \frac{1}{\sum_{i=1}^n w_i}\sum_{i=1}^n w_i\overline{f}(\mathbf{J}_i)~~\mathrm{s.t.}~~\mathbf{J}_i(\mathbf{x}) = \mathbf{I}(W(\mathbf{x};\mathcal{G}_i)), \nonumber
\end{equation}
where $\mathbf{J}_i$ is the warped image of $\mathbf{I}_i$ based on the geometry of the object $\mathcal{G}_i$ using a warping function $W$. In the subsequent sections, we describe a warping function that transforms to an egocentric image patch to an object centric canonical representation.

\subsubsection{Object Frontalization}
Consider a 3D object $\mathcal{O}$ seen from an egocentric camera. We approximate the object geometry with a cuboid where we denote its surface as $\Pi_\mathcal{O}$. Using the 3D object surface, we {\em frontalize} the egocentric image such that it is seen from the fronto parallel camera angle at a fixed distance, i.e., 
\begin{align}
    W(\mathbf{x};\mathcal{G}) = \mathbf{H}_s \mathbf{H}_\mathcal{O}
\end{align}
where $\mathbf{H}_\mathcal{O} \in \mathds{R}^{3\times 3}$ is the homography that transforms the egocentric image, i.e.,
\begin{align}
    \mathbf{H}_\mathcal{O} = \mathbf{K} \mathbf{R}_{\mathcal{O}} \mathbf{K}^{-1}
\end{align}
where $\mathbf{K}$ is the intrinsic parameter of the first-person camera, and $\mathbf{R}_{\mathcal{O}} \in SO(3)$ is the rotation matrix that accounts for the relative transform between object and camera, i.e., $\mathbf{R} = \begin{bmatrix}\mathbf{r}_x^\mathsf{T} & \mathbf{r}_y^\mathsf{T} & \mathbf{r}_z^\mathsf{T}\end{bmatrix}$ where $\mathbf{r}_z = \mathbf{n}_\mathcal{O}$ ($\mathbf{n}_\mathcal{O}$ is the surface normal of the object cuboid surface $\Pi_\mathcal{O}$), $\mathbf{r}_y = \mathbf{g}$ ($\mathbf{g}$ is the gravity direction), and $\mathbf{r}_x = \mathbf{r}_y \times \mathbf{r}_z$. This object centric frontalization (1) eliminates perspective distortion in the egocentric image; and (2) produces a canonical visual representation aligned with the gravity direction regardless of the orientation of the first-person camera.

\begin{figure}
\label{fig:frontal}\includegraphics[width=0.45\textwidth]{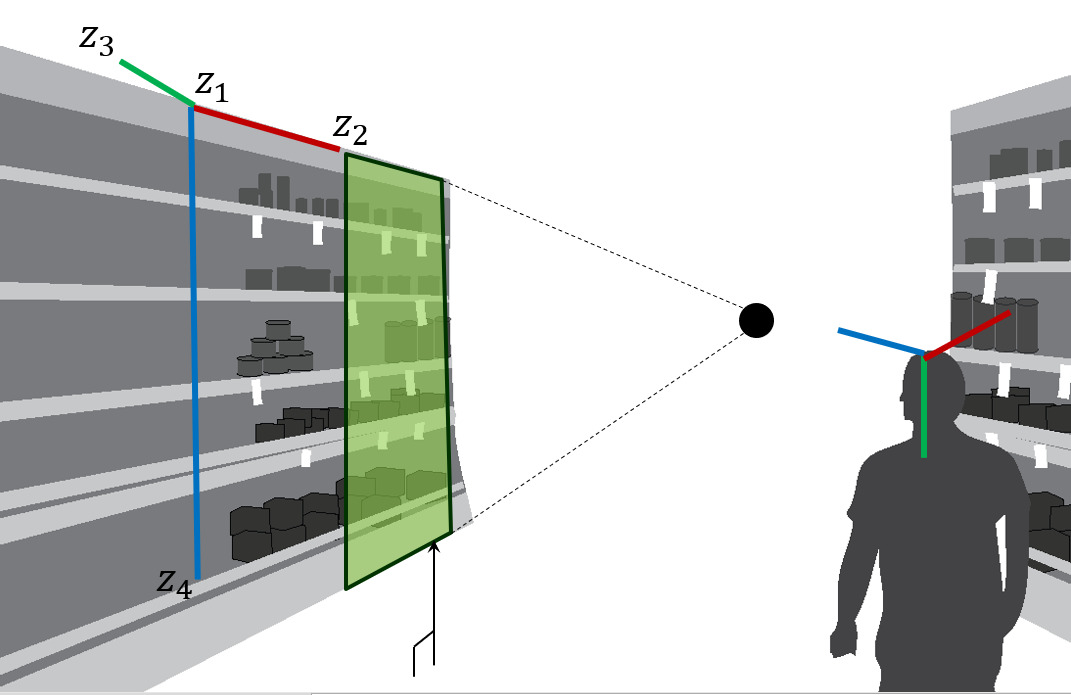}
\caption{Re-orienting to a fronto-parallel view}
\end{figure}

\begin{figure*}
    \centering
    \subfigure{\label{Fig:raw_1}\includegraphics[width=0.8cm]{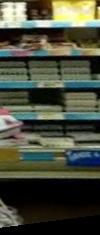}}
    \subfigure{\label{Fig:raw_1}\includegraphics[width=0.8cm]{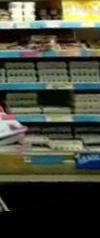}}
    \subfigure{}{\label{Fig:raw_1}\includegraphics[width=0.8cm]{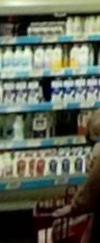}}
    \subfigure{\label{Fig:raw_1}\includegraphics[width=0.8cm]{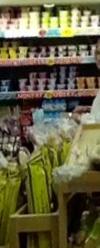}}
    \subfigure{\label{Fig:raw_1}\includegraphics[width=0.8cm]{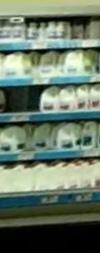}}
    \subfigure{\label{Fig:raw_1}\includegraphics[width=0.8cm]{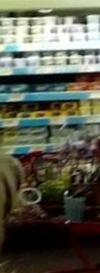}}
    \subfigure{\label{Fig:raw_1}\includegraphics[width=0.8cm]{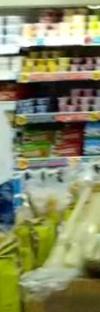}}
    \subfigure{\label{Fig:raw_1}\includegraphics[width=0.8cm]{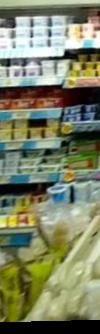}}
    \subfigure{\label{Fig:raw_1}\includegraphics[width=0.8cm]{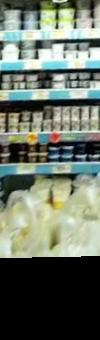}}
    \subfigure{\label{Fig:raw_1}\includegraphics[width=0.8cm]{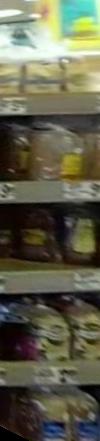}}
    \subfigure{\label{Fig:raw_1}\includegraphics[width=0.8cm]{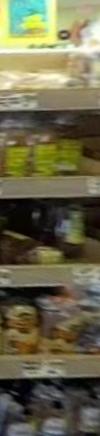}}
    \subfigure{\label{Fig:raw_1}\includegraphics[width=0.8cm]{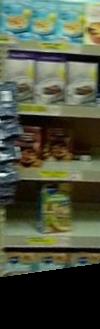}}
    \subfigure{\label{Fig:raw_1}\includegraphics[width=0.8cm]{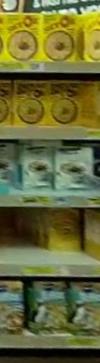}}
    \subfigure{\label{Fig:raw_1}\includegraphics[width=0.8cm]{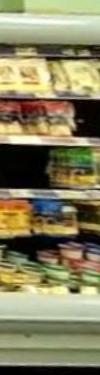}}
    \subfigure{\label{Fig:raw_1}\includegraphics[width=0.8cm]{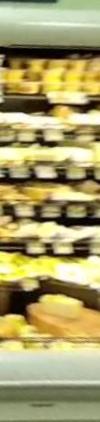}}
    \subfigure{\label{Fig:raw_1}\includegraphics[width=0.8cm]{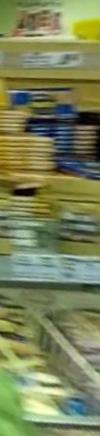}}
    \subfigure{\label{Fig:raw_1}\includegraphics[width=0.8cm]{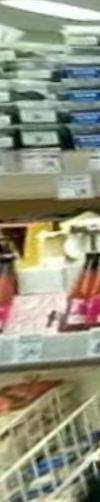}}
    \subfigure{\label{Fig:raw_1}\includegraphics[width=0.8cm]{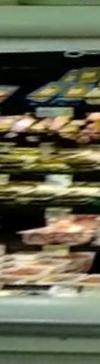}}
    \subfigure{\label{Fig:raw_1}\includegraphics[width=0.8cm]{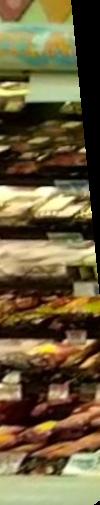}}
    \subfigure{\label{Fig:raw_1}\includegraphics[width=0.8cm]{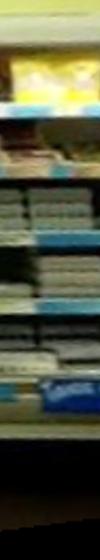}}
    \subfigure{\label{Fig:raw_1}\includegraphics[width=0.8cm]{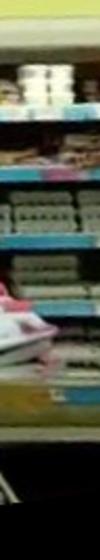}}
    \subfigure{\label{Fig:raw_1}\includegraphics[width=0.8cm]{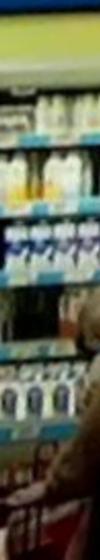}}
    \subfigure{\label{Fig:raw_1}\includegraphics[width=0.8cm]{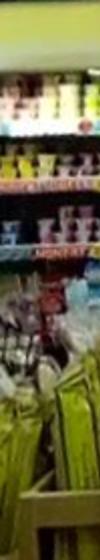}}
    \subfigure{\label{Fig:raw_1}\includegraphics[width=0.8cm]{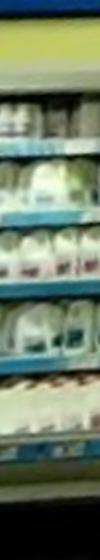}}
    \subfigure{\label{Fig:raw_1}\includegraphics[width=0.8cm]{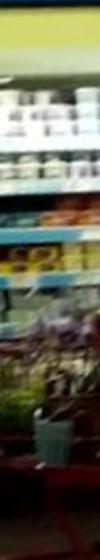}}
    \subfigure{\label{Fig:raw_1}\includegraphics[width=0.8cm]{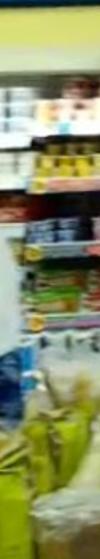}}
    \subfigure{\label{Fig:raw_1}\includegraphics[width=0.8cm]{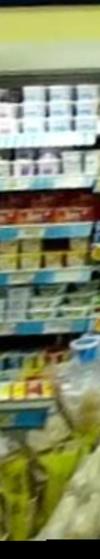}}
    \subfigure{\label{Fig:raw_1}\includegraphics[width=0.8cm]{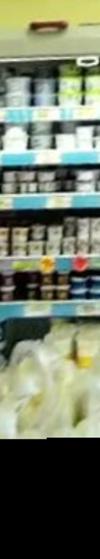}}
    \subfigure{\label{Fig:raw_1}\includegraphics[width=0.8cm]{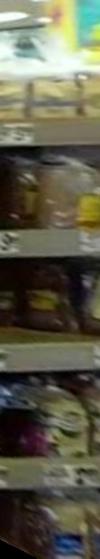}}
    \subfigure{\label{Fig:raw_1}\includegraphics[width=0.8cm]{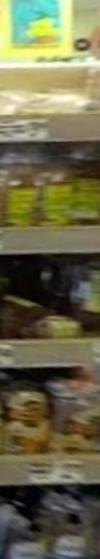}}
    \subfigure{\label{Fig:raw_1}\includegraphics[width=0.8cm]{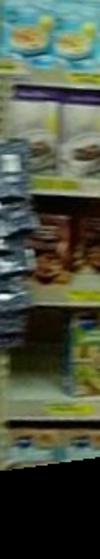}}
    \subfigure{\label{Fig:raw_1}\includegraphics[width=0.8cm]{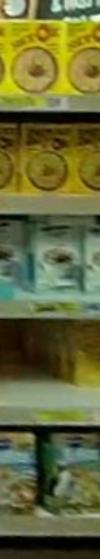}}
    \subfigure{\label{Fig:raw_1}\includegraphics[width=0.8cm]{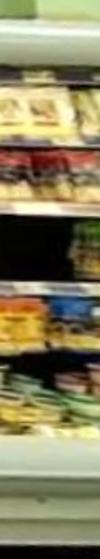}}
    \subfigure{\label{Fig:raw_1}\includegraphics[width=0.8cm]{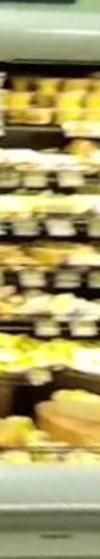}}
    \subfigure{\label{Fig:raw_1}\includegraphics[width=0.8cm]{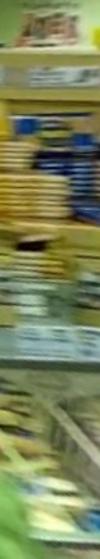}}
    \subfigure{\label{Fig:raw_1}\includegraphics[width=0.8cm]{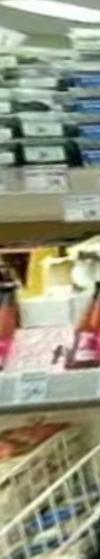}}
    \subfigure{\label{Fig:raw_1}\includegraphics[width=0.8cm]{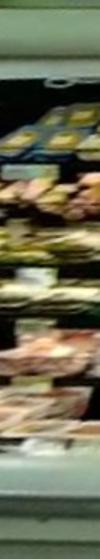}}
    \subfigure{\label{Fig:raw_1}\includegraphics[width=0.8cm]{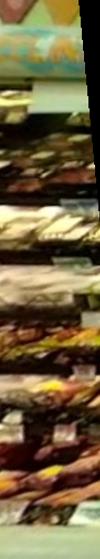}}

\caption{Image Scaling (a) top row - unscaled images, (b) bottom row - scaled images}
\label{Fig:scaling}
\vspace{-3mm}
\end{figure*}

$\mathbf{H}_\mathcal{O}$ rectifies the egocentric image with respect to the object centric coordinate, while its scale may vary depending on the depth of the object. We further account for the depth to be seen as a canonical size. Concretely, we scale the image such that the camera center translates to a fixed distance, $z=D$ where $z$ is the depth of the object and $D$ is the canonical distance. However, it is challenging because the 3D reconstruction is defined up to scale and orientation, and therefore, the depth scale is arbitrary. Instead, we leverage the scene height $H$, e.g., ceiling to the ground plane, as a reference distance to scale the image. Let $\delta v$ be pixel height of object in image (after applying $\mathbf{H_\mathcal{O}}$), and $f_y$ be vertical focal length in pixel (from the intrinsic camera parameter). Take two 3D points from a vertical slice of this object, one from top and one from bottom; then the two only differ in their $\mathbf{y}$ coordinate by $H$. Thus, in image space the two only differ in the $\mathbf{v}$-coordinate.

From the camera model, we have ${f_y H}/{z} = \delta v$ and since we want $z' = H$, we obtain,
\begin{equation}
    {z'}/{z} = {\delta v}/{f_y}\,.
\end{equation}

And so, our scaling factor for the image is $s = {f_y}/{\delta v}$. The transformation may be specified as $\mathbf{H_s} = \mathrm{diag}(s,s,1)$.

Each semantic section is thus, viewed from a canonical \emph{depth} and \emph{relative orientation}. This transformation prepares the ground for learning semantics from the scene.

\subsubsection{Extraction of semantics}
Armed with geometric stability and a dataset annotated with 3D bounding boxes, we are now able to apply the idea of reconfigurability from section~\ref{sec:reconfigurability} by extracting uniform width strips that lie within the bounding box. Since we have a frontal view of the section with the camera's $\mathbf{x}$ axis aligned with one of the axes of the section, we uniformly sample strips along the image $\mathbf{u}$ axis with the vertical pixel span given by:
\begin{equation}
    \delta v' = s  \delta v = \frac{f_y}{\delta v} \delta v = f_y
\end{equation}
Sampling along the $\mathbf{u}$ axis corresponds to sampling non-uniformly along the polar coordinate $\theta$ and we call these drawn strips $J_{\theta}$. We might attain multiple aligned views from a scene corresponding to different sections but as long as they do not overlap, we need not worry about binning/conflicts.

\subsection{Adaptation}
Warping ensures that strips, which are otherwise fragile to change in geometry, are in a format ready to be understood by semantic approaches. A purely classification based approach would work if we assume an identical distribution for test images taken in ``test stores'' as those used for training, taken in different stores. This is however false in general, given variations in illumination, clutter and image quality caused the head motion (e.g.: more motion blur, blurry images due to scale-up). To mitigate these factors, we learn an adaptation function $\mathcal{F}$ that transforms images from the test domain into training domain.

We first learn a discriminator function $\mathcal{D}$ that learns the decision boundary between strips from train stores versus test stores. The training loop for this function comprises $B$ strips randomly selected from each dataset, train and test. Let $J^{train}_{\theta}$ and $J^{test}_{\theta}$ be two such images, $\mathcal{D}(J_{\theta})$ be the probability that a strip is drawn from the test domain. Using a class label of $1$ for test domain, the optimizer minimizes the associated log loss at each step: 
\begin{equation}
    L(\mathcal{D}) = \displaystyle\sum_{i=1}^{B} - \log{\mathcal{D}(J_{\theta}^{test,i})} - \log{(1 - \mathcal{D}(J_{\theta}^{train,i}))}\,.
\end{equation}
We then learn the generative function~$\mathcal{F}$ that adapts (transforms) images from test to the training domain. That is, $\mathcal{F}(J^{test}_{\theta})$ must ``fool'' $\mathcal{D}$. We learn $\mathcal{F}$ via a residual mapping $\mathcal{R}$. More precisely, let $x_{test}$ be the features extracted from a test image~$J_{\theta}^{test}$ using a pretrained network, then:
\begin{equation}
    \mathcal{F}(J_{\theta}^{test}) = x_{test} + \mathcal{R}(x_{test})\,.
\end{equation}
In the presence of large amounts of data or perfect similarity between train and test stores, the residual would be zero (or very close to zero) because no transformation is needed. 

Learning a transformation by solely minimizing the classification loss from the discriminator can destroy the semantic content of the image and thus all images are mapped to either the same category or, even worse, the same image. To counter this, we learn an inverse function $\mathcal{G}$ whose job is to reconstruct image features prior to adaptation. Adding a reconstruction penalty, the combined loss function for the problem becomes:
\begin{align}
    L(\mathcal{F}, \mathcal{G}) &= - \log{(1 - \mathcal{D}(\mathcal{F}(x_{test})))} \nonumber\\
    &+ \alpha \lvert\lvert \mathcal{G}(\mathcal{F}(x_{test})) - x_{test} \rvert\rvert
\end{align}
where $\alpha$ is a hyperparameter that controls the weight of the reconstruction loss.

The adaptation module described is generic and can be fitted on top of any pretrained network. Finally, we connect a logits layer at the end of such a network and train it to classify the adapted features by minimizing the cross-entropy loss.

\subsection{Scene Representation}
Combining the different parts, the scene representation for images from a test store can be expressed as:

\begin{equation}
    f(\mathbf{I}) = \frac{1}{\sum_{i=1}^n w_i}\sum_{i=1}^n w_i\mathcal{F}(\overline{f}( \mathbf{I}(W(\mathbf{x};\mathcal{G}_i)))), \nonumber
    \label{eqn:adapted_rep}
\end{equation}

\begin{figure*}[t]
    \centering
    
    \subfigure{\label{Fig:human_pck}\includegraphics[width=0.24\textwidth]{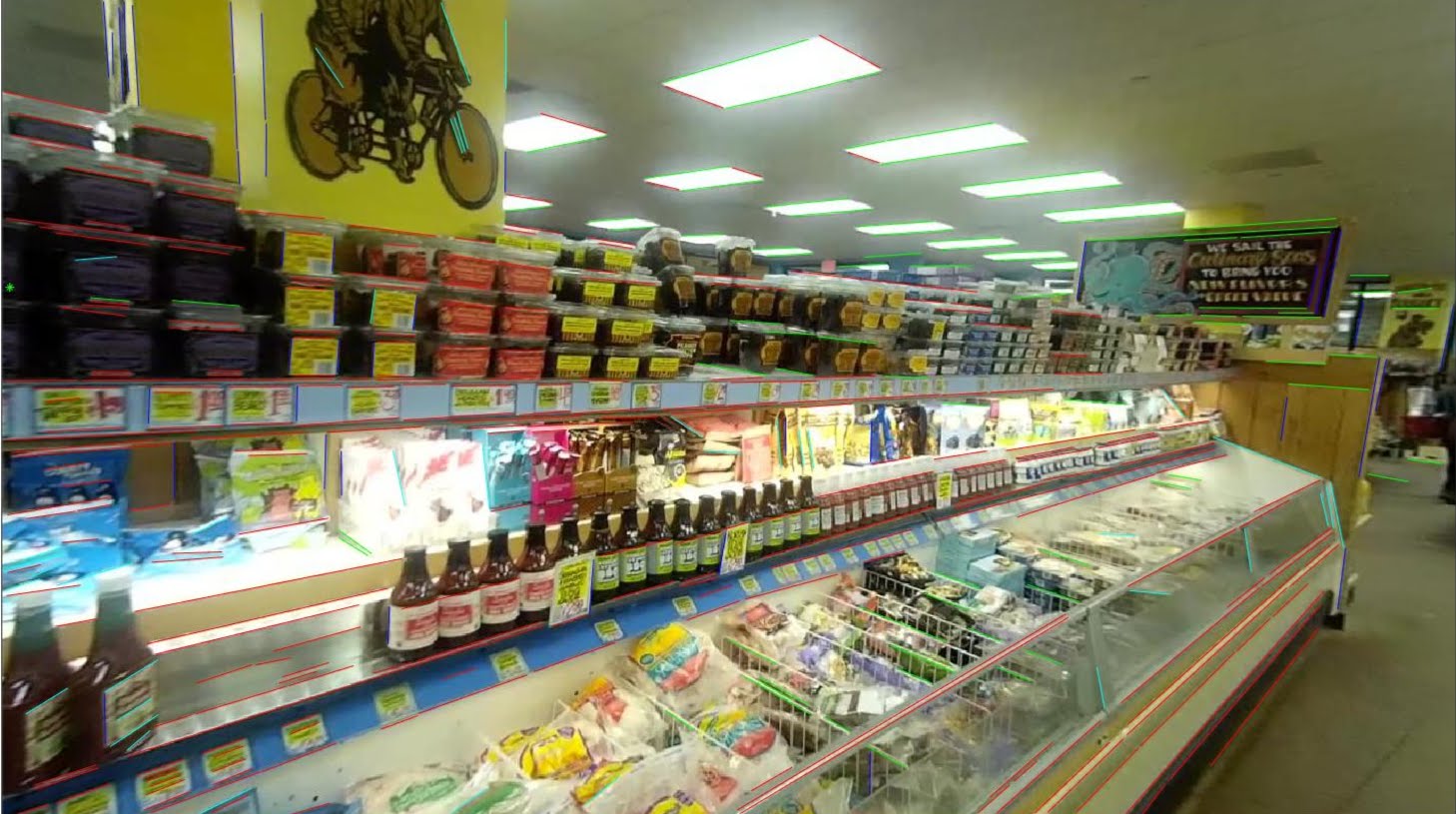}}
    \subfigure{\label{Fig:monkey_pck}\includegraphics[width=0.24\textwidth]{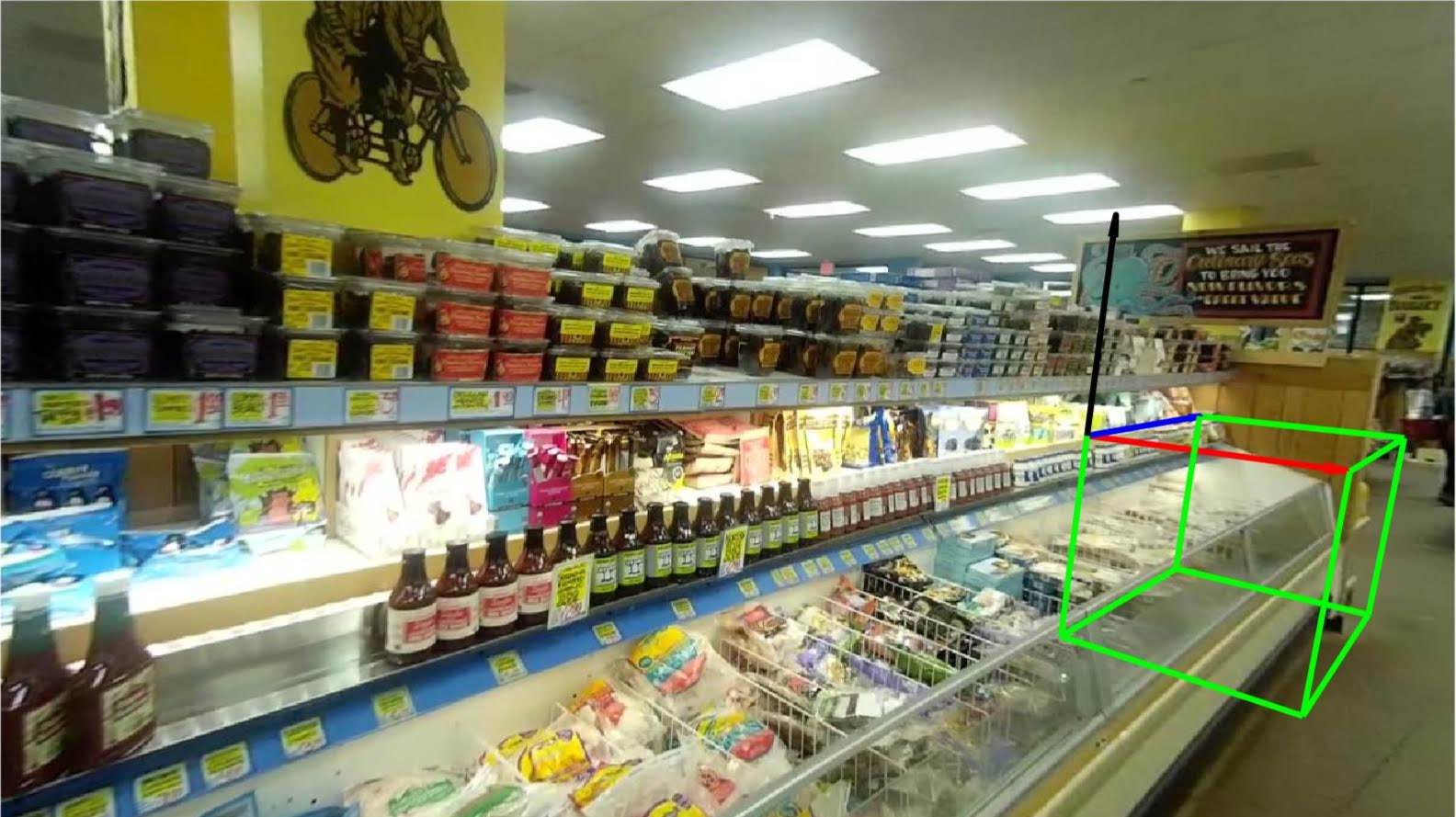}}
    \subfigure{\label{Fig:monkey_pck}\includegraphics[width=0.24\textwidth]{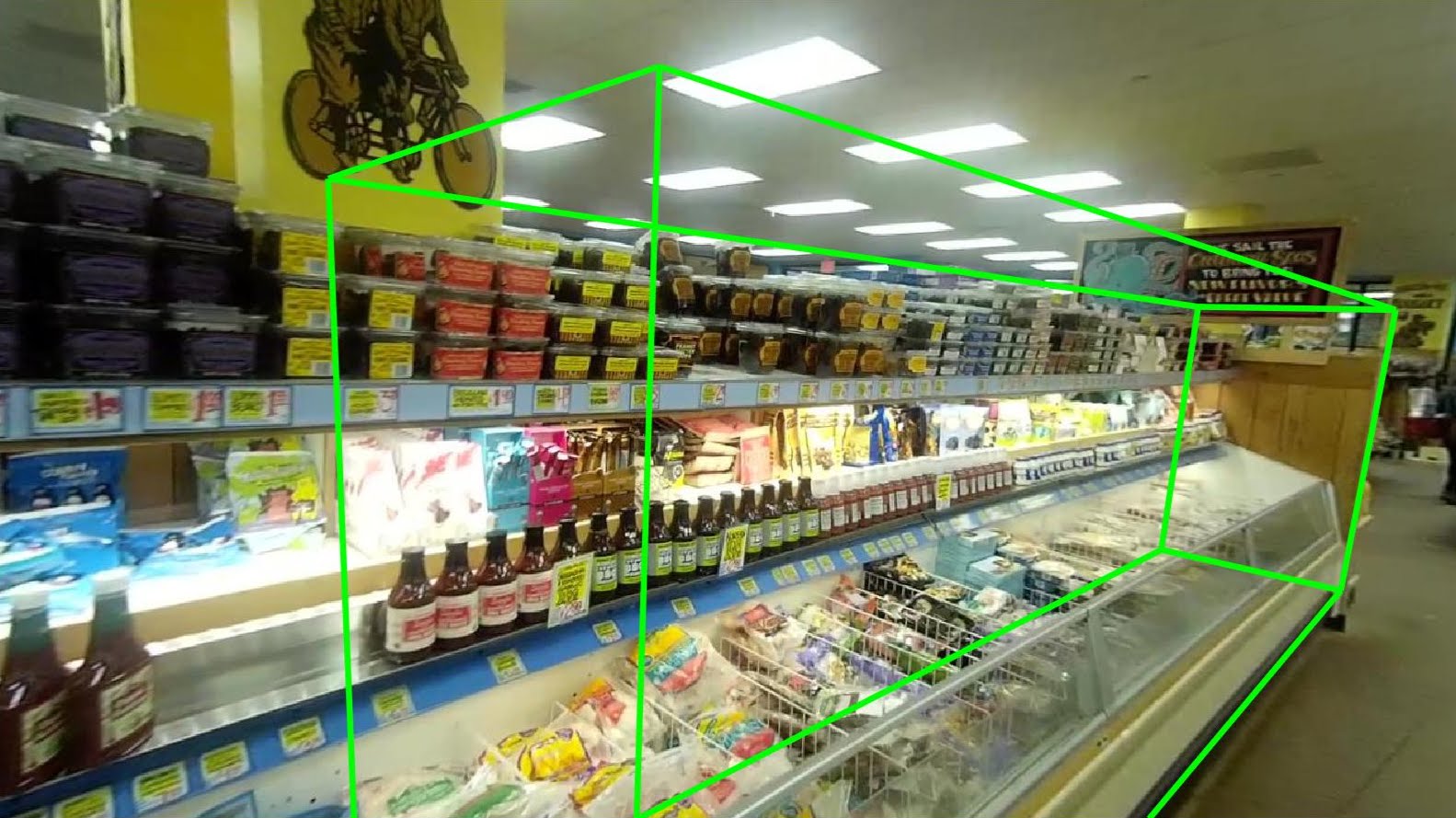}}
    \subfigure{\label{Fig:monkey_pck}\includegraphics[width=0.24\textwidth]{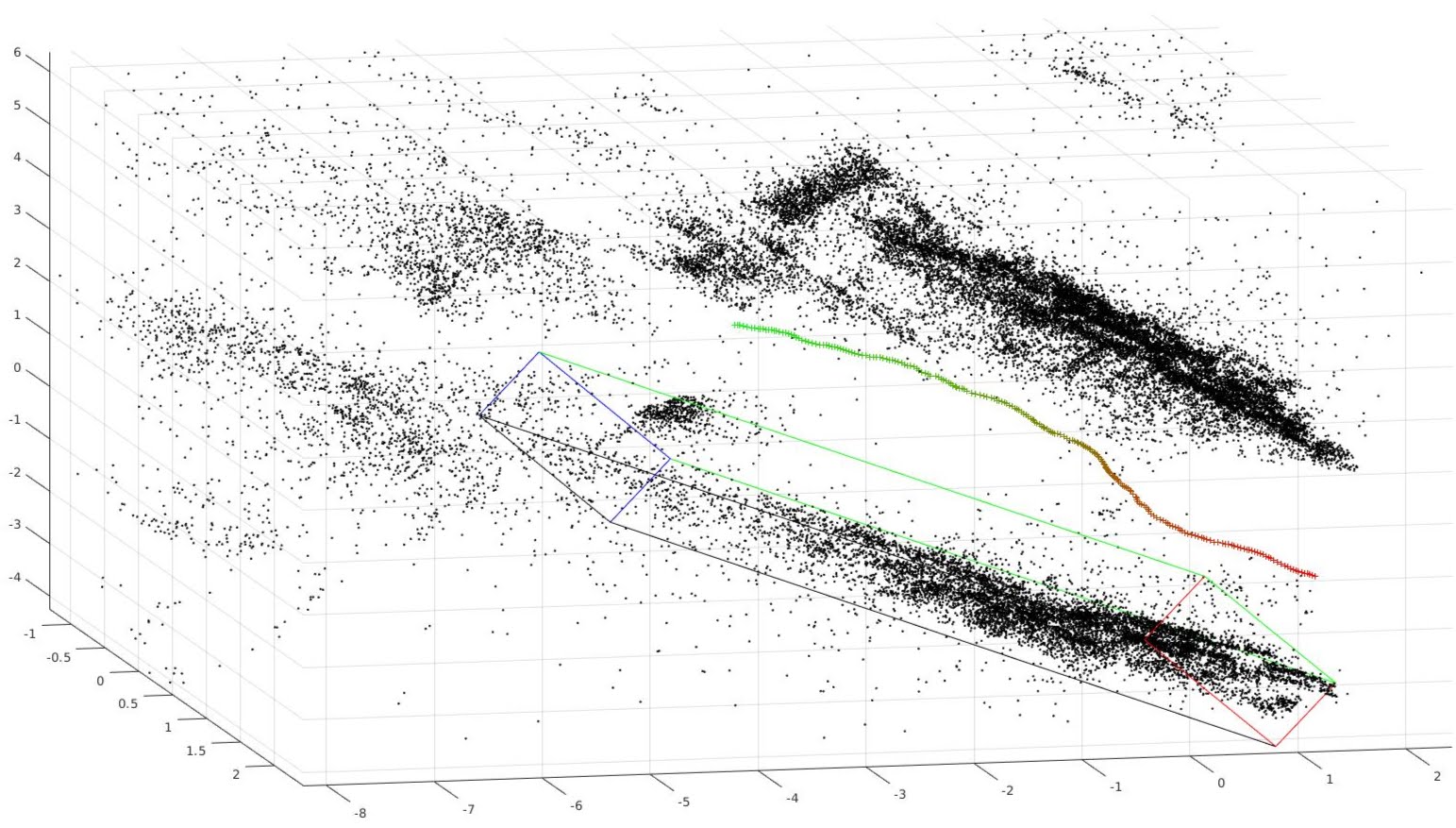}}
\caption{Annotating dataset (a) vanishing points computation, (b) projecting box aligned with orthogonal directions, (c) labeling section by manipulating box and (d) 3D fitted box.}
\label{fig:labeling}
\vspace{-3mm}
\end{figure*}

\section{Dataset}
We collect first-person footage from four Trader Joe's, 3 Target and 2 Whole Food stores in the local metro area using commercially available zshade glasses. The total footage collected is 90 minutes in length. For each of the videos, we extract frames at 30 fps, and reconstruct the camera trajectory and 3D space using a standard structure from motion pipeline.

\subsection{Annotation}
We label a total of 25 semantic categories using 3D bounding boxes in 30000 undistorted images~\cite{devernay:2001}. To achieve this scale, we designed a novel interface by leveraging scene geometry and reconstructed camera motion.

We assume that sections are aligned with the three principal orthogonal directions of the scene. First, we calculate three mutually orthogonal vanishing points using Hedau et al.'s~\cite{hedau:2009} algorithm. In practice, the sections inform the calculation of these points. We manually select points $vp_x$ and $vp_y$ in $X$ and $Y$ directions in camera coordinate system. Consider the 3D point where the $X$ axis (in camera coordinates) meets the projective line corresponding to $vp_x$ - let this point be $X_p$. Using $K$ for the camera intrinsic parameters, $R$ and $C$ for pose, we must have
\begin{align*}
    \lambda vp_x &= K R (X_p - C)\implies X_p - C = \lambda R^T K^{-1} vp_x\,.
\end{align*}
The $X$ axis direction in 3D is thus given by $\text{unit}(X_p - C) = \text{unit}(R^T K^{-1} vp_x)$. Similarly, the $Y$ direction and the gravity vector is obtained as the cross product of the two. We observe that using the cross product gives more stable gravity vector compared to simply using the output from the algorithm.

Once we have principal axes, we triangulate an origin point in 3D using pixel correspondence between two images. Using the origin and axes, we construct a bounding box in 3D that is projected onto the image. Keyboard input to the interface can be used to move each of the faces of the box in the normal direction. Since the bounding box's corners are specified in 3D, these labels are easily propagated to the rest of the images in the reconstructed batch. Thus, 1 image labels 200 others. Figure~\ref{fig:labeling} illustrates steps in the labeling process.

\section{Experiments}

\subsection{Setup}
Post-scaling, we extract strips from rectified section views by sampling at a uniform width of 100 pixels. These strip images are then re-sized to a height of 500px whilst preserving aspect ratio and then centered onto a 500x500 black background. Throughout our experiments, we use an off-the-shelf pre-trained network~\cite{kaiming:2015} as a feature extractor. Our training set consists of strips from two stores and we test models on the unseen third store. We consider 6 semantic labels---bread, cereal, cheese, dairy, frozen-food and meat. The only preprocessing applied is RGB-mean subtraction.

\subsection{Quantitative Results}

\subsubsection{Domain Adaptation}
Our discriminator $\mathcal{D}$ is a fully-connected neural network with 2048 units--- same as the length of feature vectors. Similarly, the residual function $\mathcal{R}$ for adaptation and reconstruction function $\mathcal{G}$ are fully-connected 2048 units neural networks. Weight decay used in training $\mathcal{R}$ is $1$ and $\alpha$ is chosen to be $1$.

\begin{figure}
    \centering
    \includegraphics[width=0.5\textwidth]{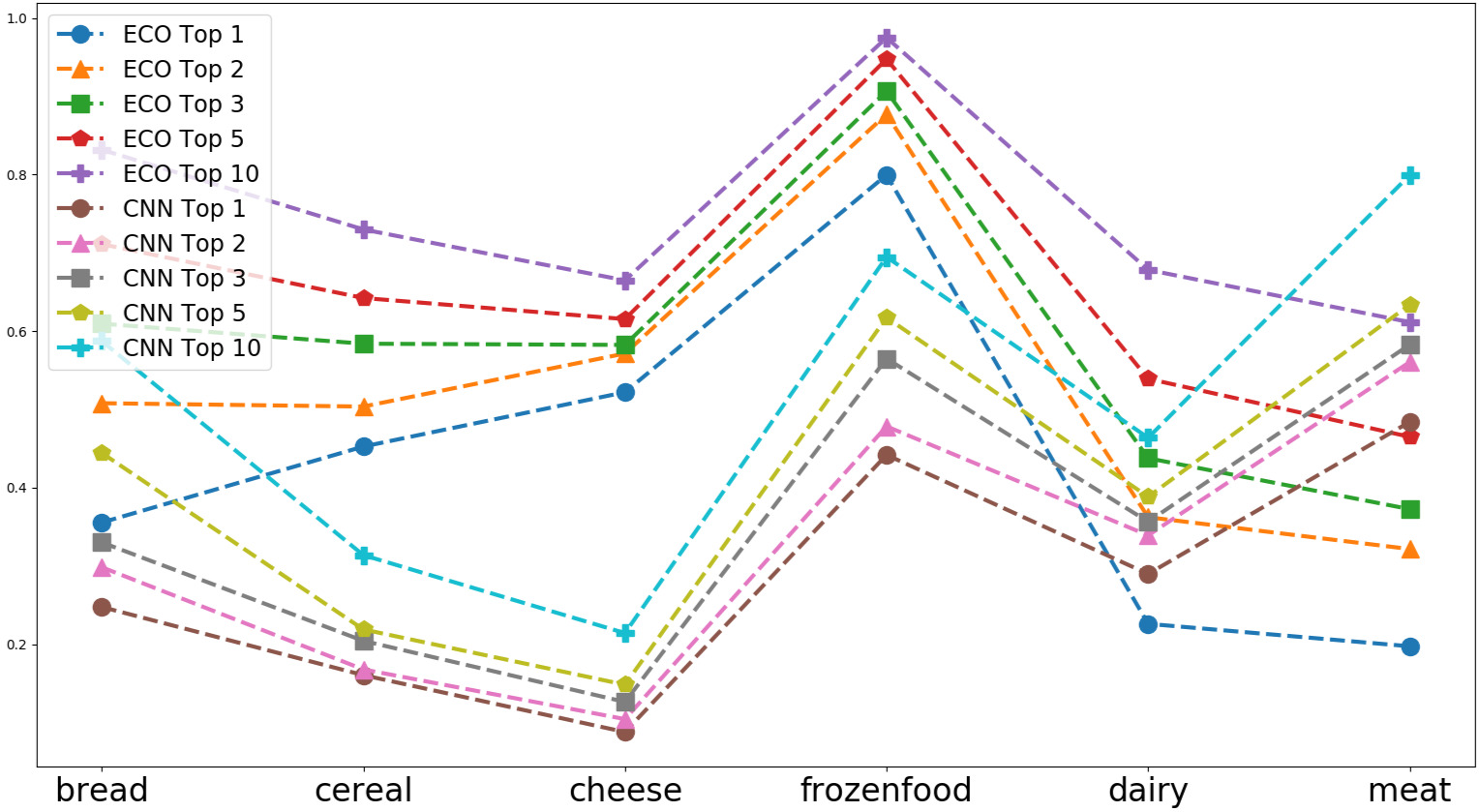}
    \caption{Recall curve on nearest neighbor classification for our Adapted features vs CNN}
    \label{fig:recall_curve}
\end{figure}

We evaluate the discriminative properties of our scene descriptor from Eqn~\ref{eqn:adapted_rep}  ($w_i = 1 \; \forall \; i$) against global CNN descriptors on a classification task. There are a total of 3124 scenes in the training dataset, and 1660 scenes in the test dataset. To demonstrate the expressiveness of our features, we use nearest neighbor based classification. As borne out by the recall curve in Figure \ref{fig:recall_curve}, our features consistently outperform global CNN features on all categories barring meat.

\subsubsection{Classification}
To demonstrate that our compositional strips can be used to learn semantics in a transferrable way, we train a logistic classifier on strips from our train store and test on the third. The setup comprises a logits layer connected to the last layer of ResNet-152~\cite{kaiming:2015}; an L2 regularization term plus softmax cross-entropy loss are minimized. There are a total of 16136 strips in training, and 8478 in test. Results in Table \ref{table:strip_classification} clearly show that our induced warping facilitates learning. 

\begin{table}
\footnotesize
\setlength{\tabcolsep}{5pt}
  \centering
  \begin{tabular}{ | l | c | c | c | c | c | c |}
    \hline
    \textbf{Category} & Bread & Cereal & Cheese & Dairy & Frozen-food & Meat \\  \hline
    \textbf{Accuracy} & 0.36 & 0.51 & 0.39 & 0.6 & 0.49 & 0.33 \\    \hline
  \end{tabular}
  \vspace{0.25mm}
  \caption{Strip classification on Store 3}
  \label{table:strip_classification}
\end{table}

\subsection{Qualitative Results}
\begin{figure}
    \centering
    \subfigure{\label{fig:bread_nbr}\includegraphics[width=0.32\columnwidth]{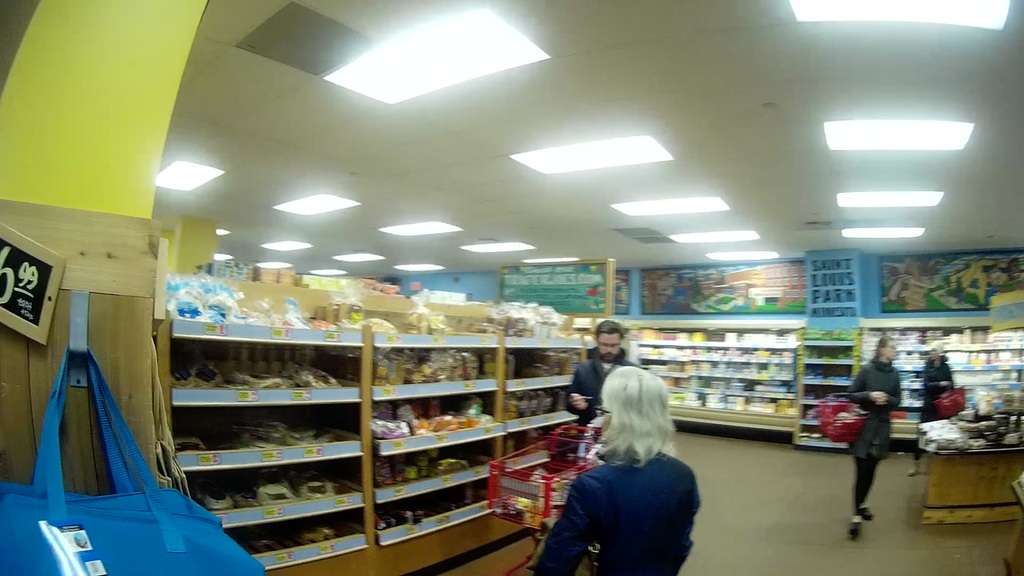}}
    \vspace{-3mm}
    
    \subfigure{\label{Fig:monkey_pck}\includegraphics[width=0.32\columnwidth]{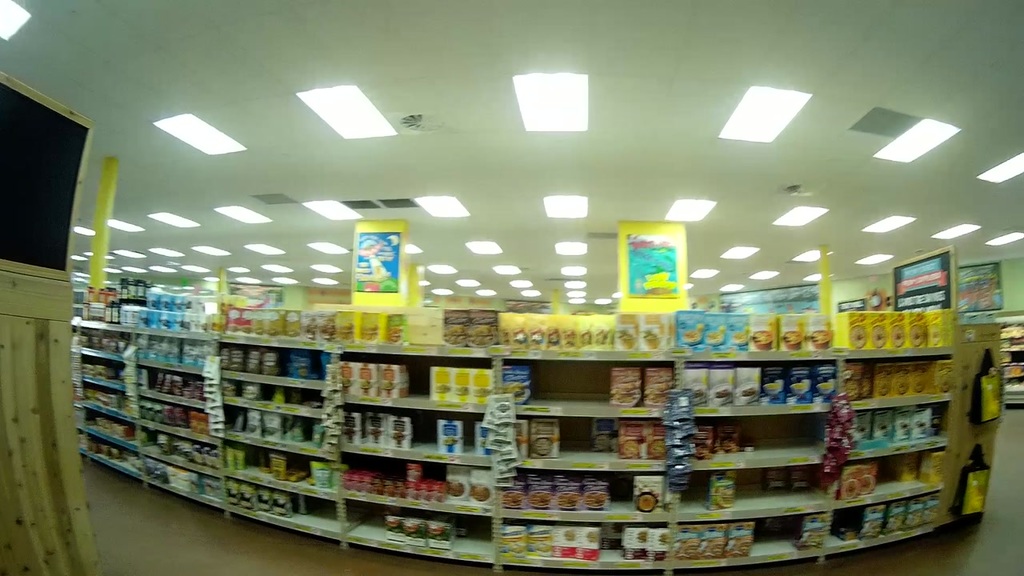}}
    \subfigure{\label{Fig:monkey_pck}\includegraphics[width=0.32\columnwidth]{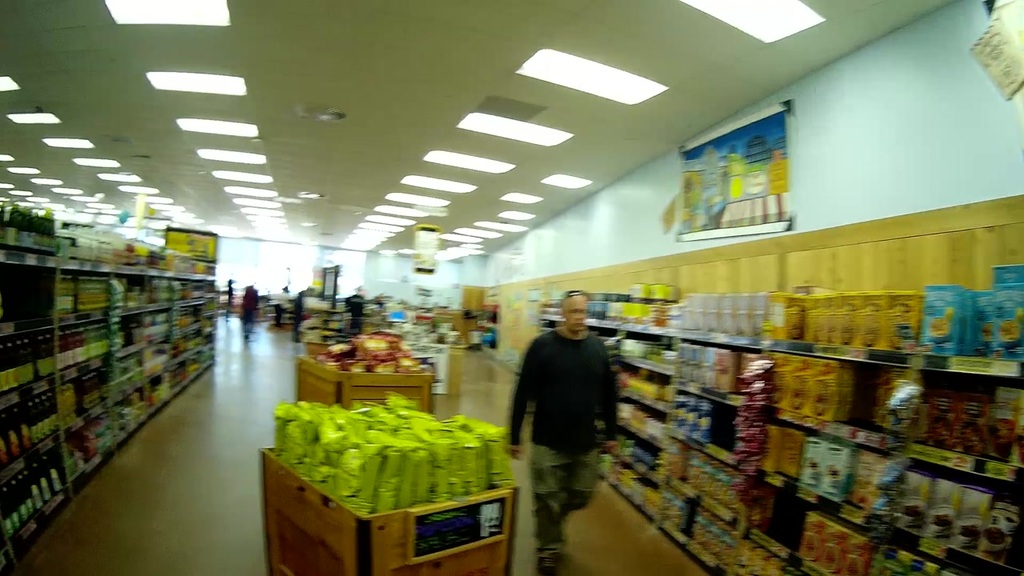}}
    \subfigure{\label{Fig:monkey_pck}\includegraphics[width=0.32\columnwidth]{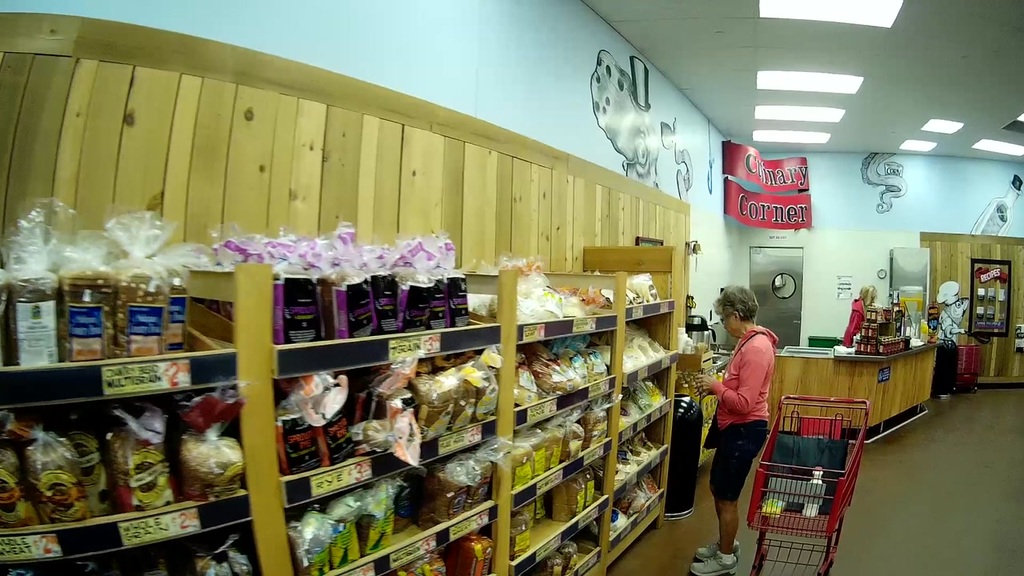}}
    \vspace{-3mm}
    
    \subfigure{\label{Fig:monkey_pck}\includegraphics[width=0.32\columnwidth]{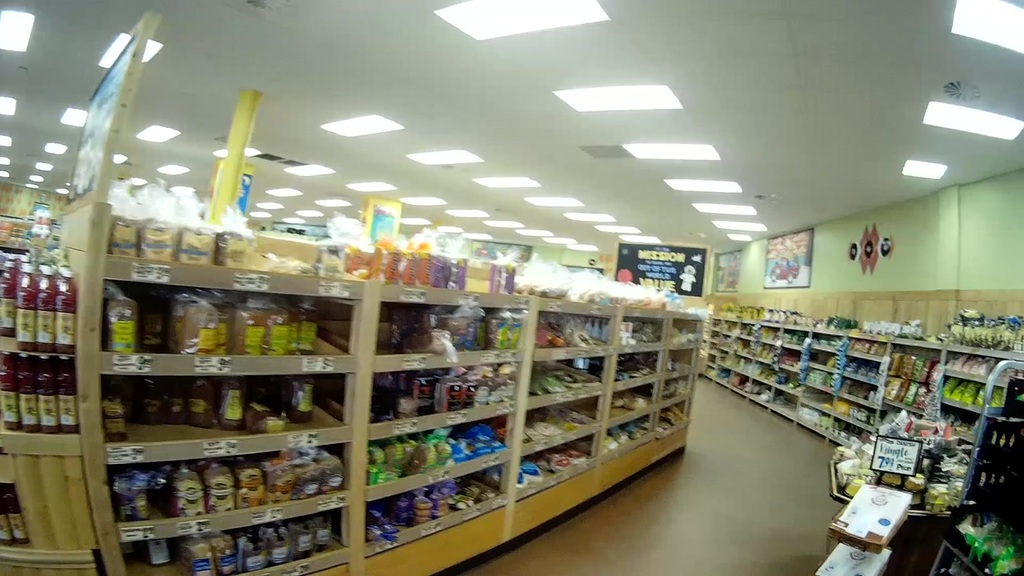}}
    \subfigure{\label{Fig:monkey_pck}\includegraphics[width=0.32\columnwidth]{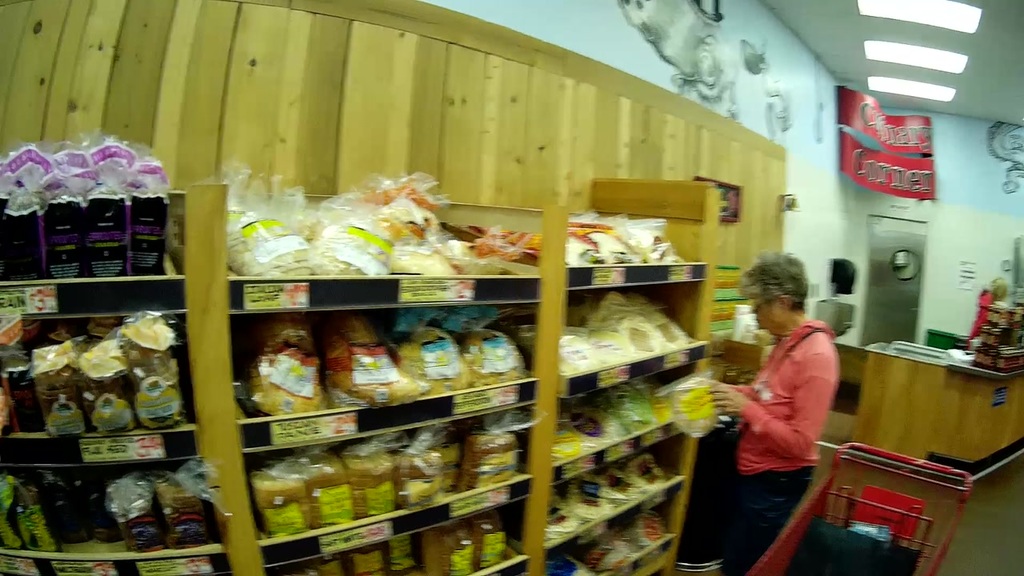}}
    \subfigure{\label{Fig:monkey_pck}\includegraphics[width=0.32\columnwidth]{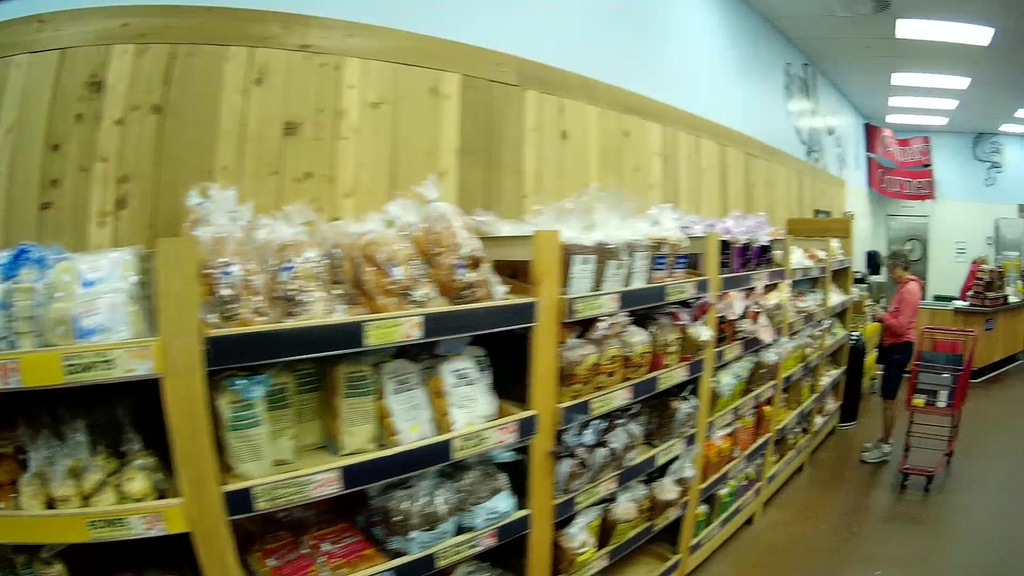}}
    \caption{Top: bread query, middle: CNN, bottom: ECO}
\label{fig:nearest_nbr}
\vspace{-1mm}
\end{figure}

\begin{figure}
    \centering
    \subfigure{\label{Fig:human_pck}\includegraphics[width=0.32\columnwidth]{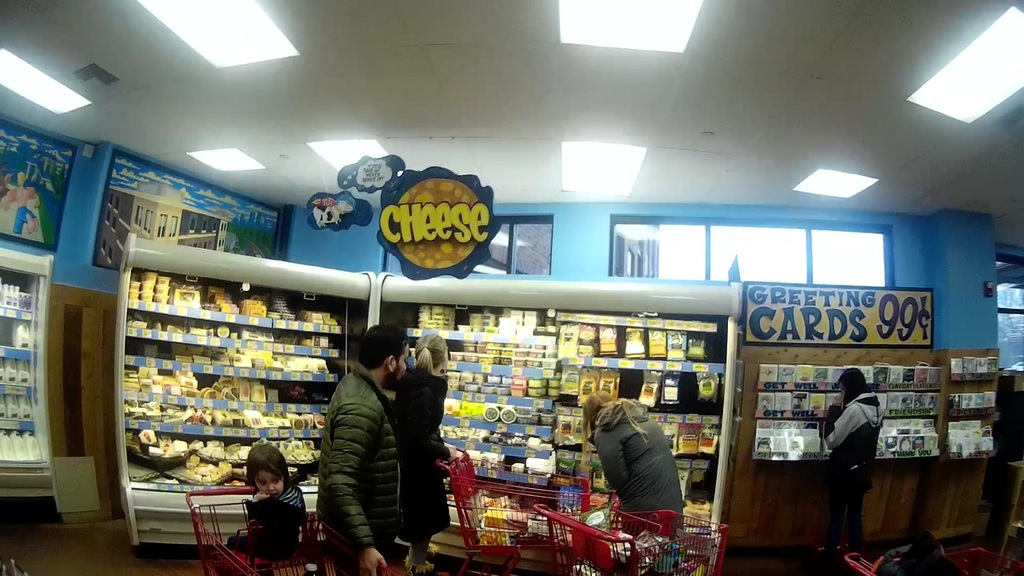}}
        \vspace{-3mm}
    
    \subfigure{\label{fig:cheese_nbr}\includegraphics[width=0.32\columnwidth]{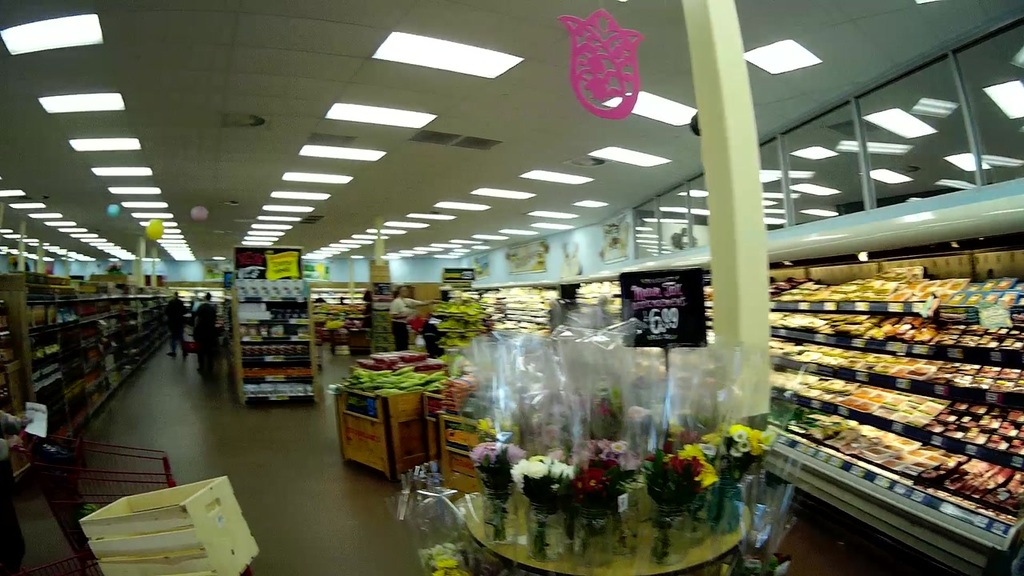}}
    \subfigure{\label{Fig:monkey_pck}\includegraphics[width=0.32\columnwidth]{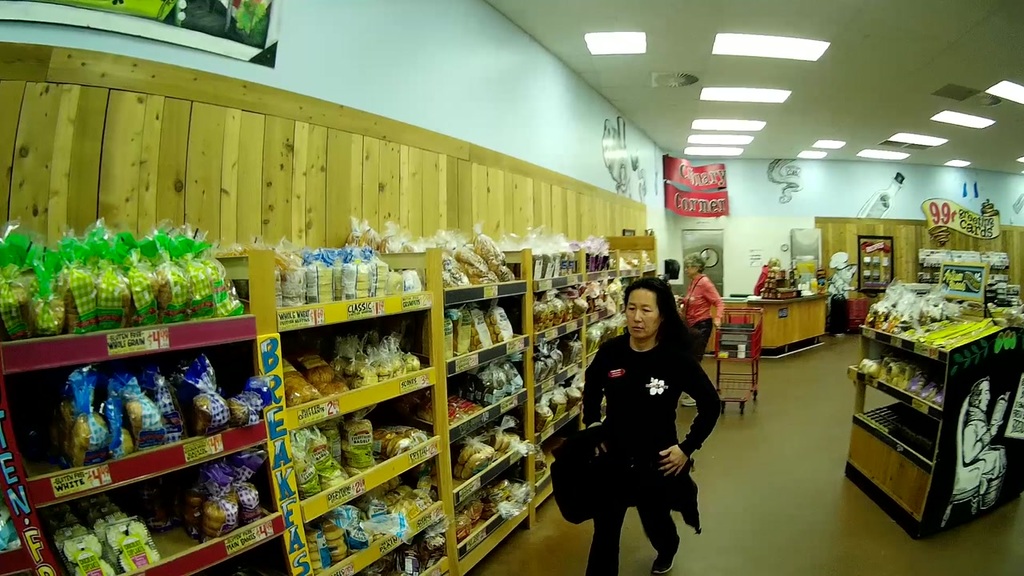}}
    \subfigure{\label{Fig:monkey_pck}\includegraphics[width=0.32\columnwidth]{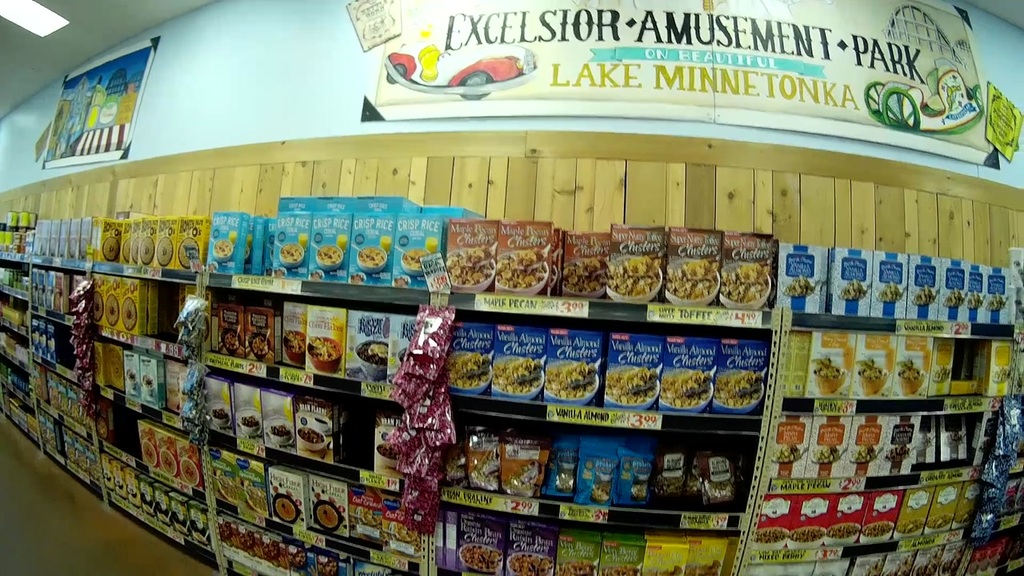}}
        \vspace{-3mm}

    \subfigure{\label{Fig:monkey_pck}\includegraphics[width=0.32\columnwidth]{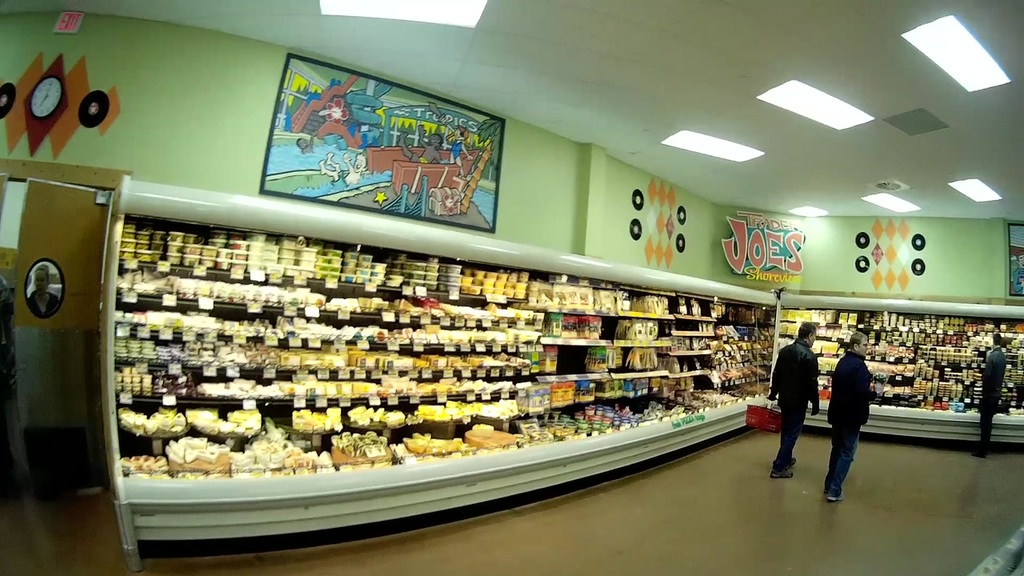}}
    \subfigure{\label{Fig:monkey_pck}\includegraphics[width=0.32\columnwidth]{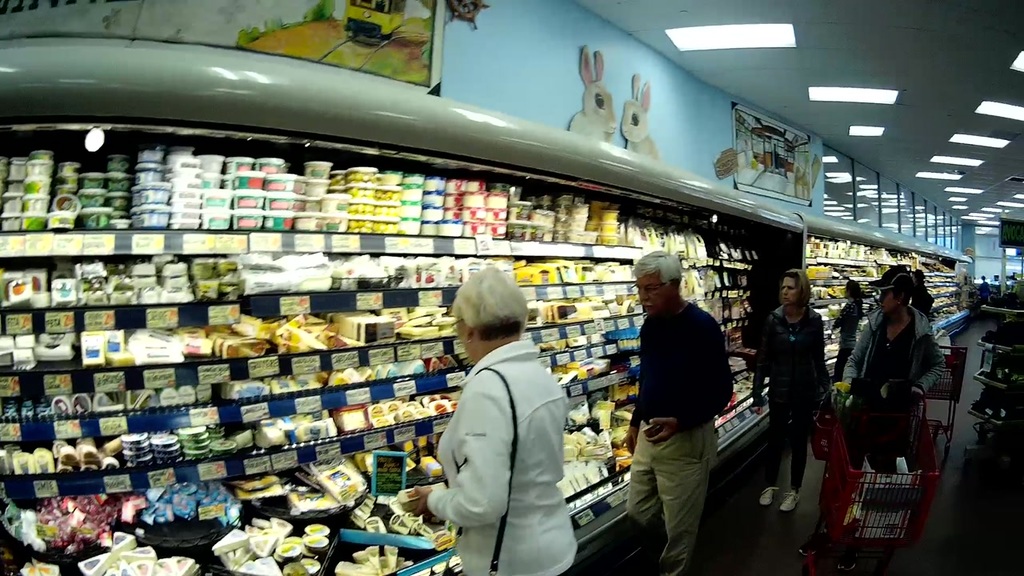}}
    \subfigure{\label{Fig:monkey_pck}\includegraphics[width=0.32\columnwidth]{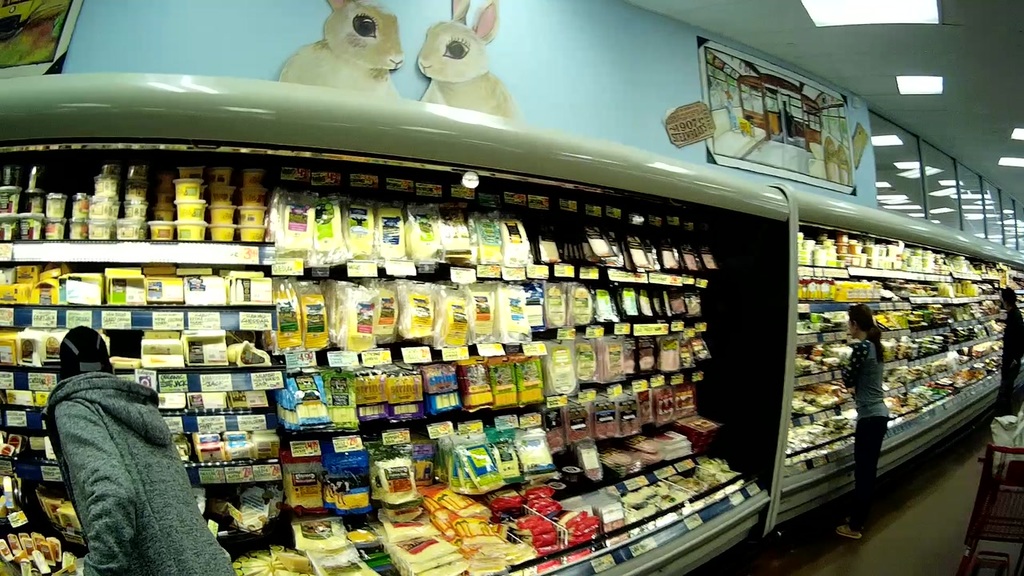}}
        \caption{Top: cheese query, middle: CNN, bottom: ECO}
    \label{fig:nearest_nbr}
\vspace{-1mm}
\end{figure}

\begin{figure}
    \centering
        
    \subfigure{\label{fig:dairy_nbr}\includegraphics[width=0.32\columnwidth]{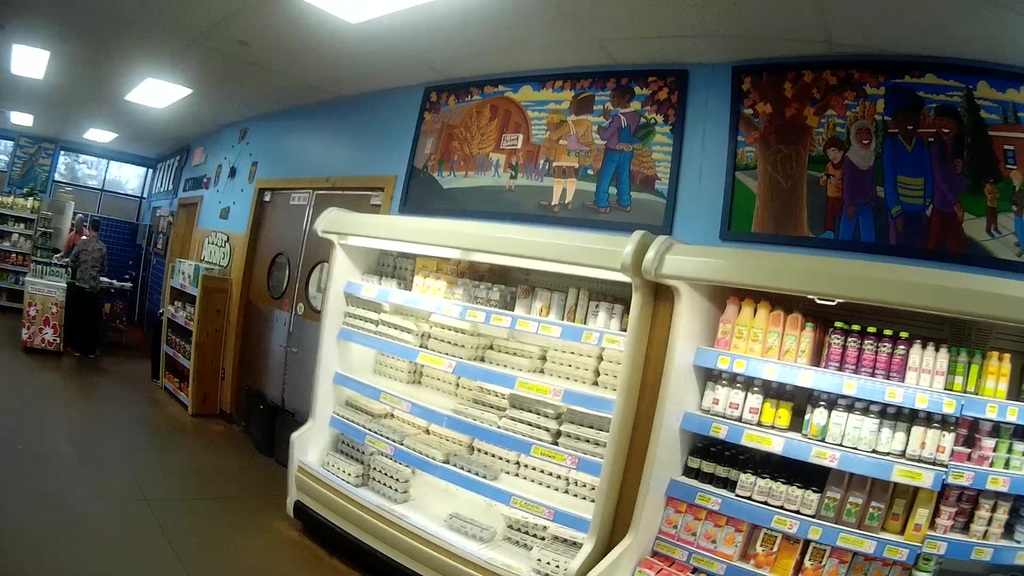}}
        \vspace{-3mm}
    
    \subfigure{\label{Fig:monkey_pck}\includegraphics[width=0.32\columnwidth]{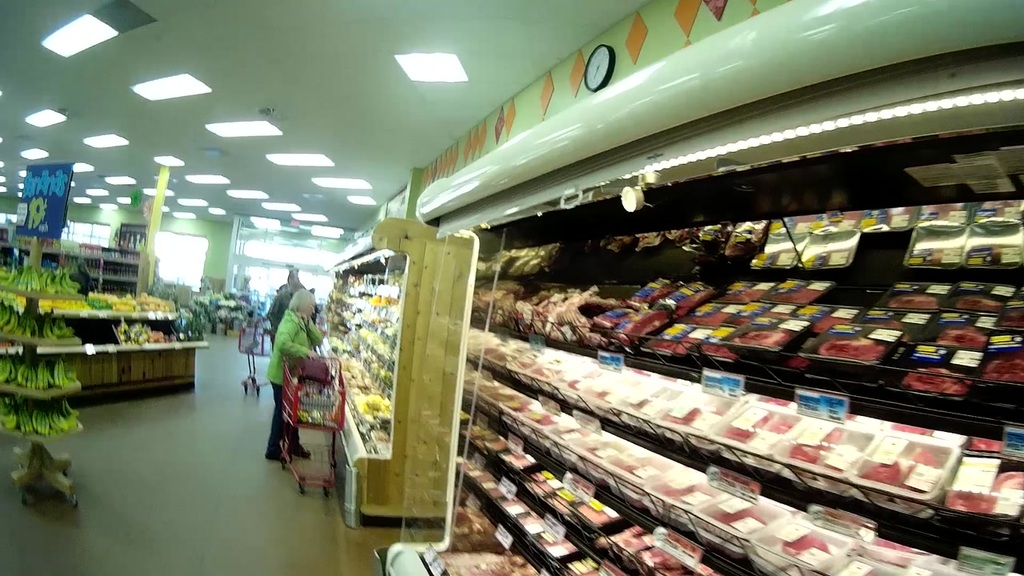}}
    \subfigure{\label{Fig:monkey_pck}\includegraphics[width=0.32\columnwidth]{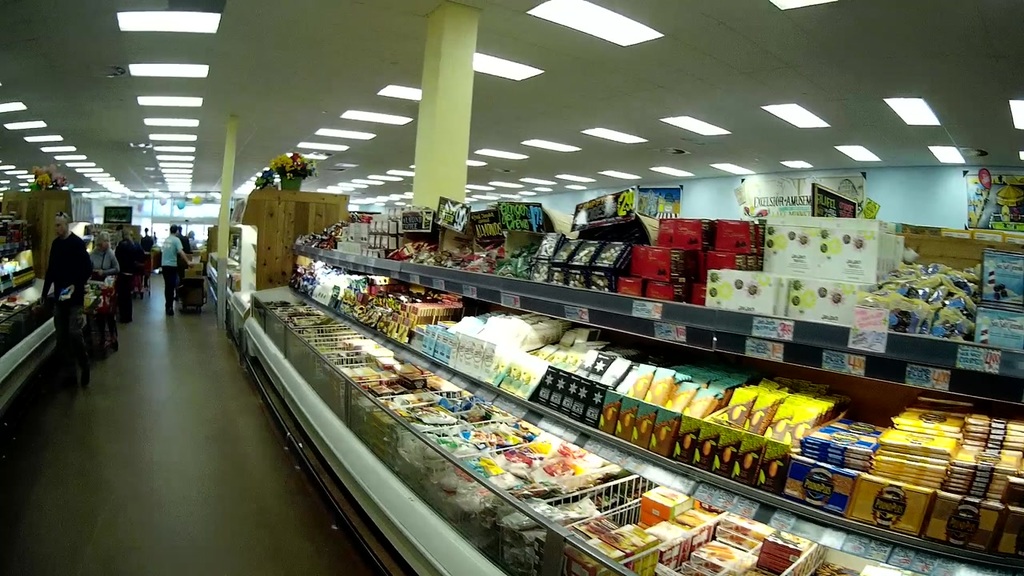}}
    \subfigure{\label{Fig:monkey_pck}\includegraphics[width=0.32\columnwidth]{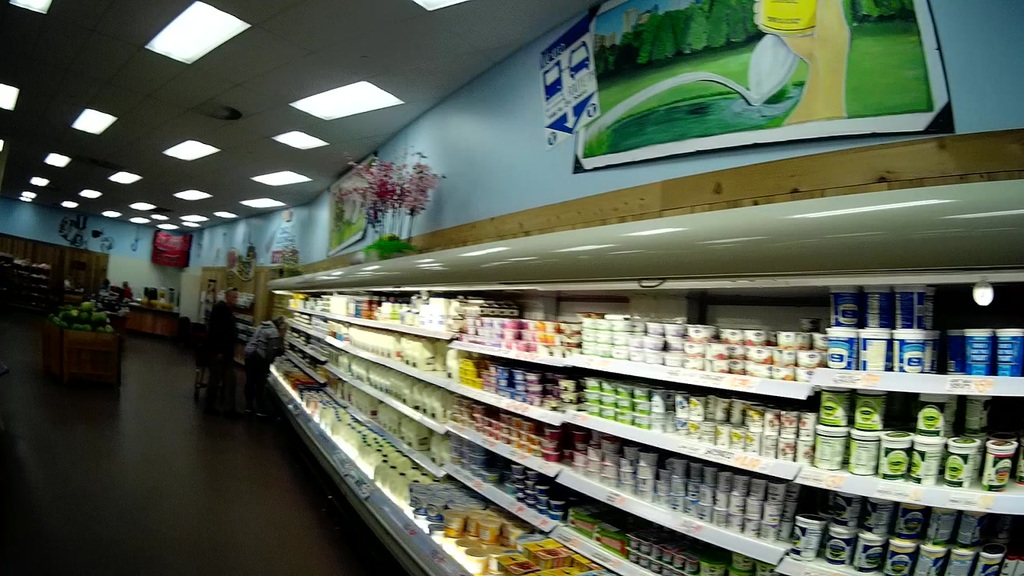}}
        \vspace{-3mm}

    \subfigure{\label{Fig:monkey_pck}\includegraphics[width=0.32\columnwidth]{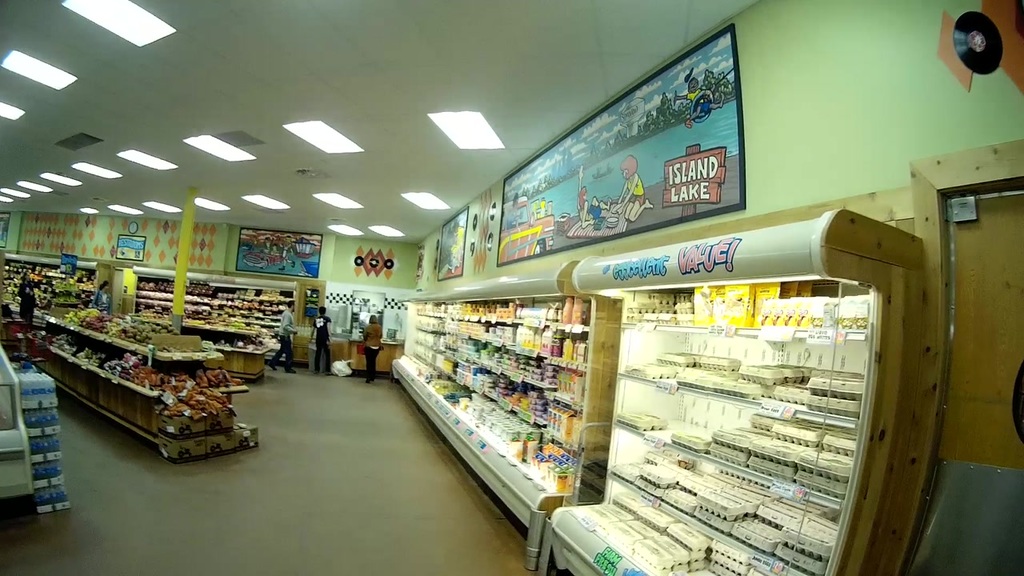}}
    \subfigure{\label{Fig:monkey_pck}\includegraphics[width=0.32\columnwidth]{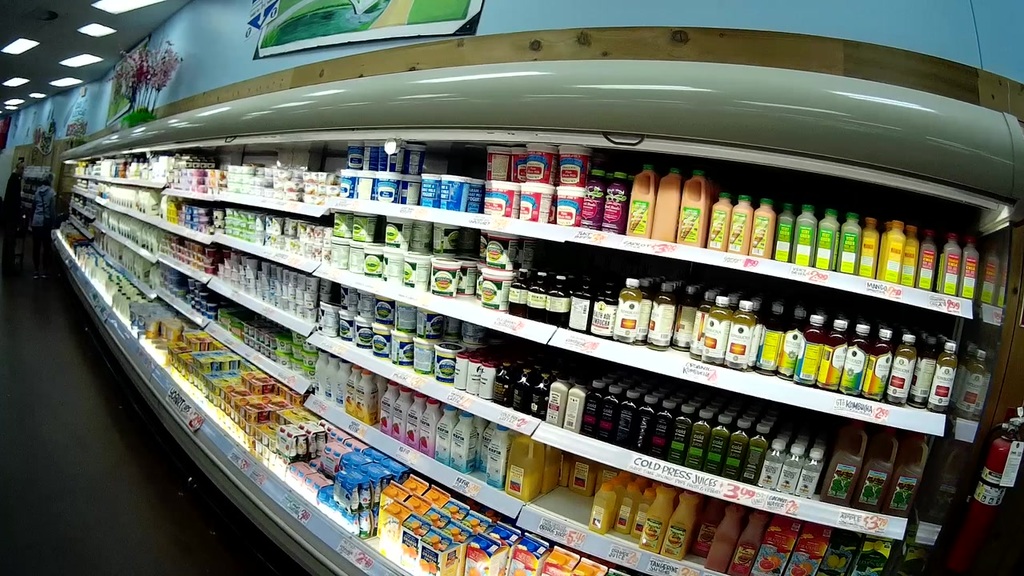}}
    \subfigure{\label{Fig:monkey_pck}\includegraphics[width=0.32\columnwidth]{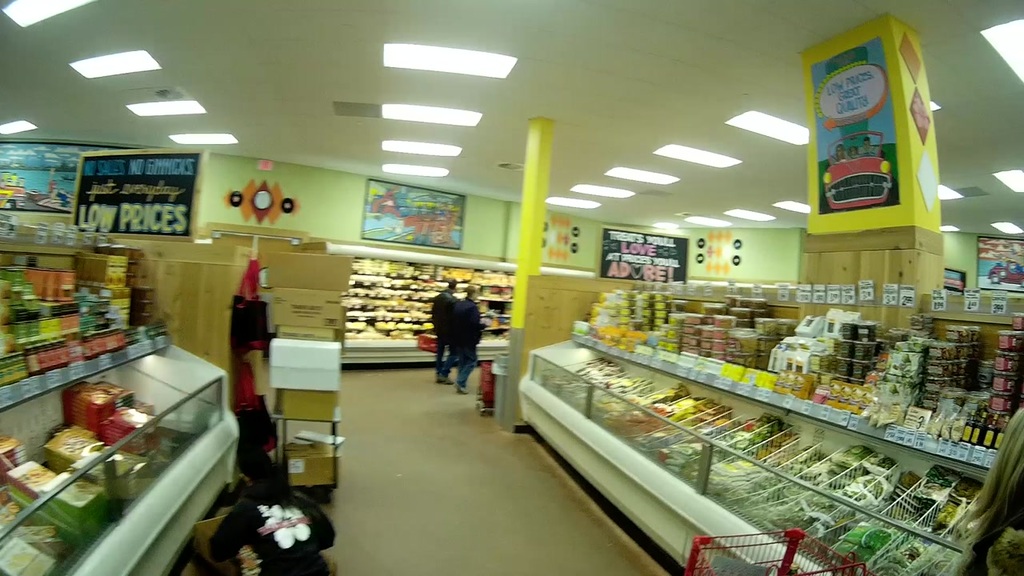}}
        \caption{Top: dairy query, middle: CNN, bottom: ECO}
\vspace{-1mm}
\end{figure}

\begin{figure}
    \centering
    \subfigure{\label{Fig:human_pck}\includegraphics[width=0.32\columnwidth]{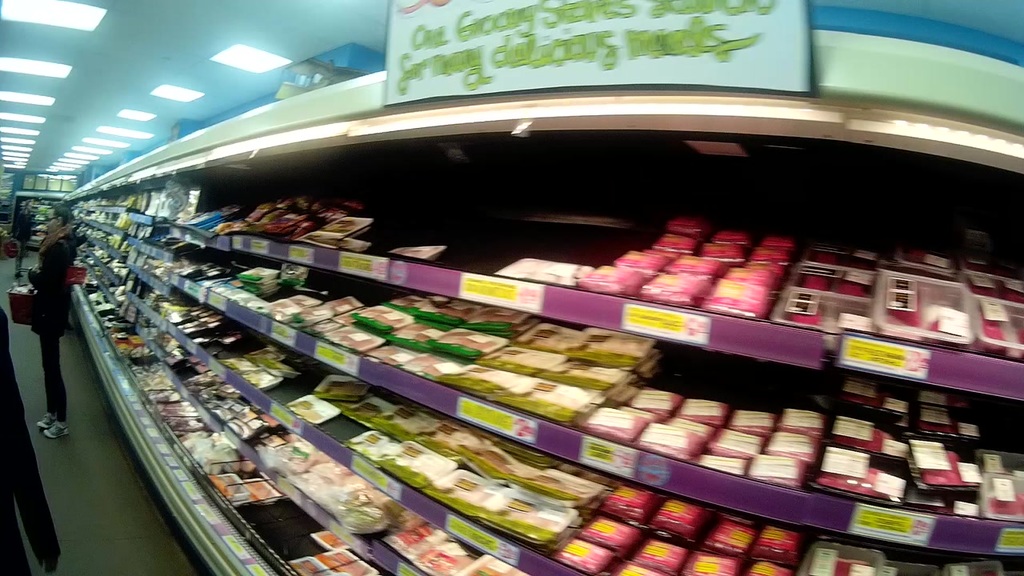}}
    \vspace{-3mm}
    
    \subfigure{\label{fig:meat_nbr}\includegraphics[width=0.32\columnwidth]{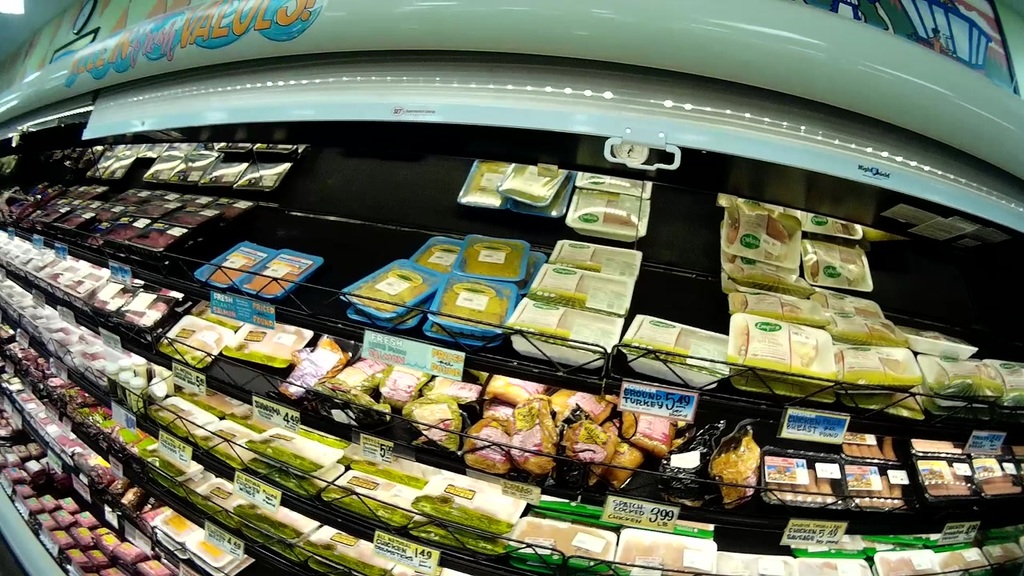}}
    \subfigure{\label{Fig:monkey_pck}\includegraphics[width=0.32\columnwidth]{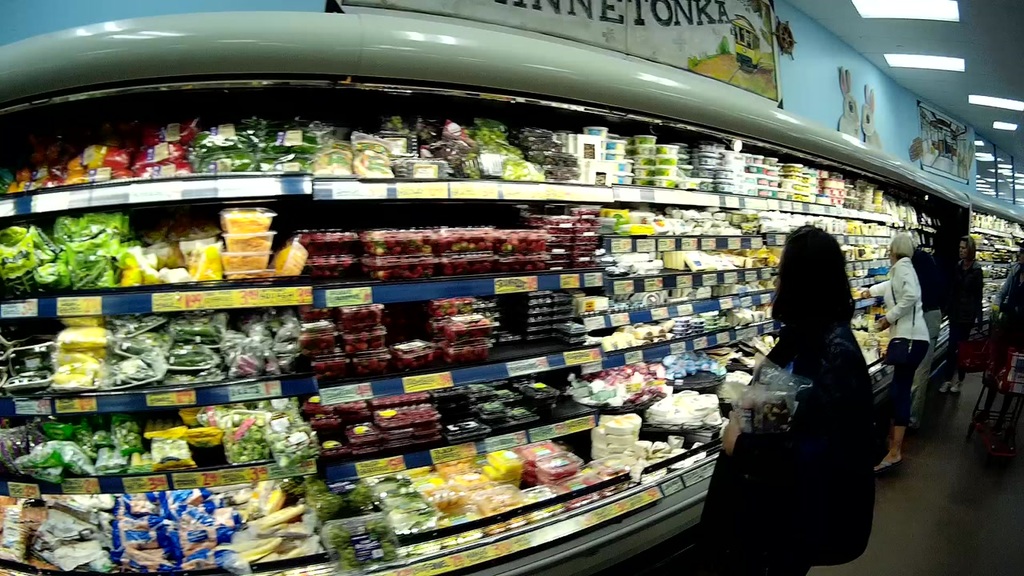}}
    \subfigure{\label{Fig:monkey_pck}\includegraphics[width=0.32\columnwidth]{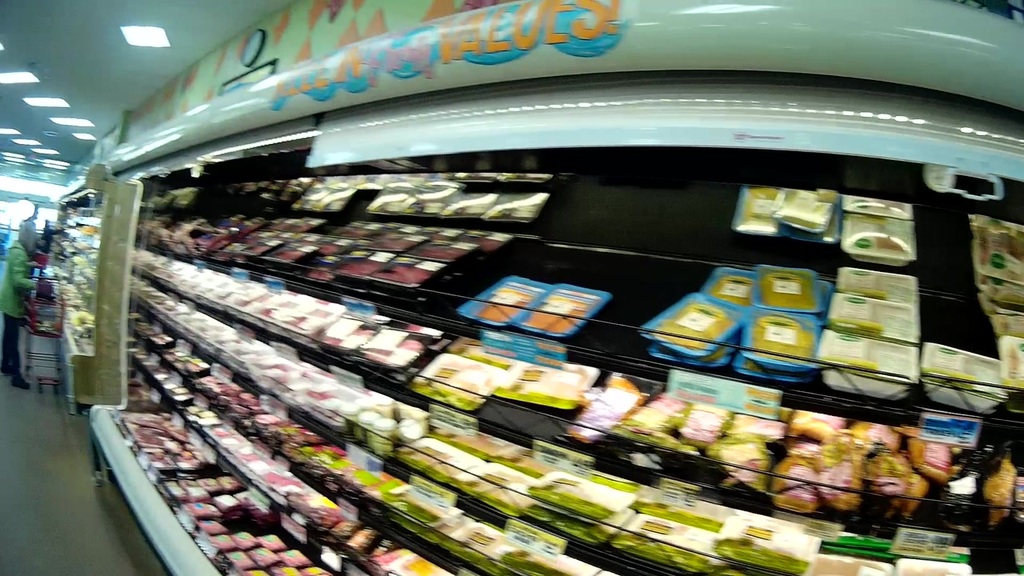}}
    \vspace{-3mm}
    
    \subfigure{\label{Fig:monkey_pck}\includegraphics[width=0.32\columnwidth]{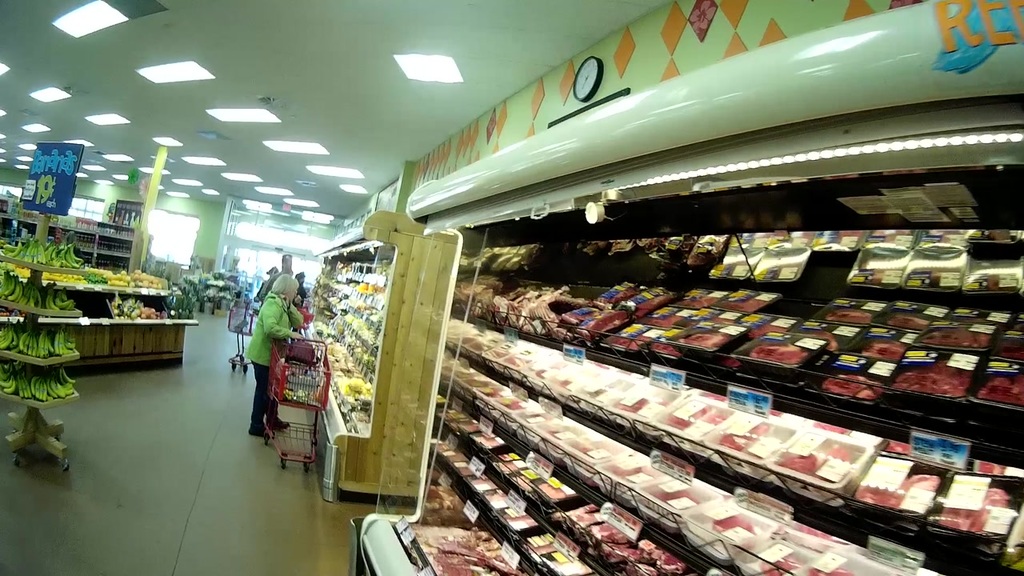}}
    \subfigure{\label{Fig:monkey_pck}\includegraphics[width=0.32\columnwidth]{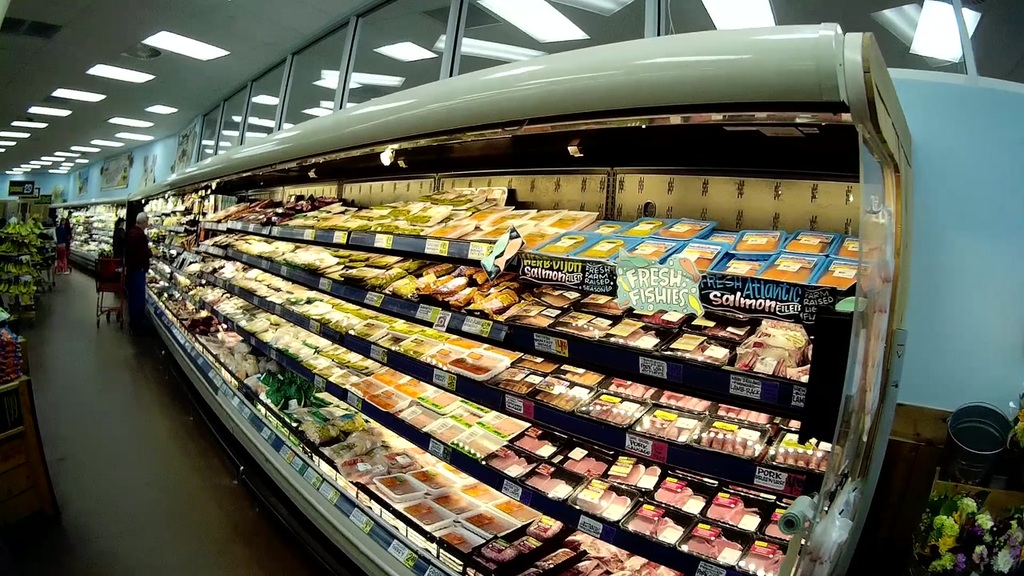}}
    \subfigure{\label{Fig:monkey_pck}\includegraphics[width=0.32\columnwidth]{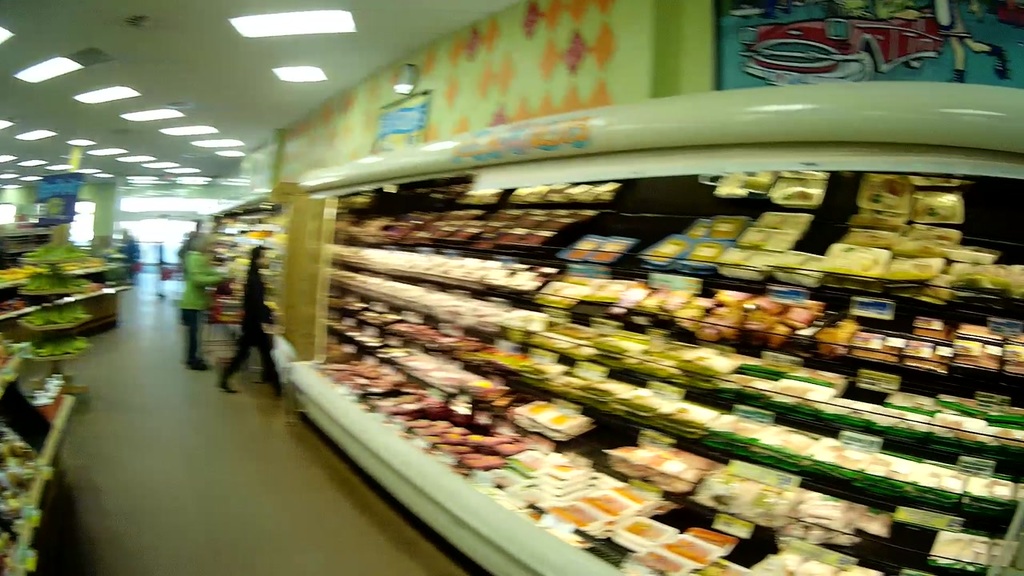}}

    \caption{Top: meat query, middle: CNN, bottom: ECO}
\label{fig:nearest_nbr}
\vspace{-1mm}
\end{figure}

We present a qualitative comparison between global CNN features and our adapted features using some sample query images for each semantic class in Figure~\ref{fig:nearest_nbr}. The top three nearest neighbors are presented after removing similar images. While CNN features routinely generate neighbors similar in background and geometry, for our features we observe neighbors drawn from diverse orientations, object proximity and very importantly, stores. The bread (\ref{fig:bread_nbr}) neighbors and query image have completely different depths; cheese (\ref{fig:cheese_nbr}) neighbors contain varying prominent parts (whole section/spread/slices); dairy (\ref{fig:dairy_nbr}) neighbors match either eggs or milk-based beverages while query contains both; meat (\ref{fig:meat_nbr}) neighbors are invariant to proximity unlike CNN neighbors.

\section{Summary}
In this paper, we propose a new representation, called ECO (Egocentric Cognitive Mapping) for higher level scene understanding that possesses the key properties of a cognitive map. It is compositional in nature, and we define an object-centric warping of scenes from canonical depth to derive the atomic patches of understanding.
We pose the problem of inferencing semantics in a new grocery store as a domain adaptation problem and construct a learning framework for the task. We demonstrate the quantitative and qualitative superiority of our ECO representation over global CNN descriptors for first-person scenes.

{\small
\bibliographystyle{ieee}
\bibliography{egbib,hs_bib,js_bib}
}

\end{document}